%% file: root.tex
\documentclass[letterpaper, 10 pt, conference, twoside]{ieeeconf}  

\IEEEoverridecommandlockouts                              

\title{\LARGE \bf
Deployment of Reliable Visual Inertial Odometry Approaches for Unmanned Aerial Vehicles in Real-world Environment
}

\author{%
  Jan Bedn\'{a}\v{r}$^{1}$, Mat\v{e}j Petrl\'{i}k$^{1}$, Kelen Cristiane Teixeira Vivaldini$^{1,2}$, Martin Saska$^{1}$
  \thanks{$^1$Jan Bedn\'{a}\v{r}, Mat\v{e}j Petrl\'{i}k, Kelen Cristiane Teixeira Vivaldini and Martin Saska are with the Faculty of Electrical Engineering, Czech Technical University in Prague, Czech Republic, {\tt\small \{jan.bednar14|matej.petrlik|martin.saska\} @fel.cvut.cz}.}
 \thanks{$^2$Kelen Cristiane Teixeira Vivaldini is with the Federal University of São Carlos, Department of Computer, Brazil, {\tt\small vivaldini@ufscar.br}.}
 }

\usepackage{xcolor}
\usepackage{graphicx}
\usepackage{adjustbox}
\usepackage{tikz}

\usepackage{mathtools}
\usepackage{leading}
\usepackage{siunitx}
\usepackage{multirow}
\usepackage{booktabs}
\usepackage{caption}
\usepackage{subcaption}
\usepackage[super]{nth}
\usepackage{balance}
\usepackage{fancyhdr}
\usepackage{hyperref} 
\hypersetup{
  colorlinks,
  citecolor=black,
  filecolor=black,
  linkcolor=black,
  urlcolor=black,
  pdfauthor={},
  pdfsubject={},
  pdftitle={}
}

\usepackage{pgfplots}
\usetikzlibrary{backgrounds,arrows,automata,shapes,positioning,calc,through}
\pgfdeclarelayer{background}
\pgfdeclarelayer{foreground}
\pgfsetlayers{background,main,foreground}

\tikzset{
  state/.style={
    rectangle,
    draw=black, very thick,
    minimum height=1.0em,
    text centered,
  },
  final_state/.style={
    rectangle,
    rounded corners,
    draw=black, very thick,
    minimum height=2em,
    text centered,
  },
  initial_state/.style={
    rectangle,
    double=white,
    double distance=1pt,
    inner sep=2pt,
    draw=black, very thick,
    minimum height=2em,
    text centered,
  },
  point/.style={
    circle,
    inner sep=0pt,
    minimum size=3pt,
    fill=red
  },
  adder/.style={
    circle,
    inner sep=2pt,
    minimum size=0.3in,
    draw=black, very thick,
    text centered
  }
}

\definecolor{dark_red}{rgb}{0.4, 0.0, 0.0}
\definecolor{light_red}{rgb}{0.8, 0.1, 0.1}
\definecolor{dark_green}{rgb}{0.0, 0.4, 0.0}
\definecolor{light_green}{rgb}{0.1, 0.8, 0.1}
\definecolor{dark_blue}{rgb}{0.0, 0.0, 0.4}
\definecolor{light_blue}{rgb}{0.1, 0.1, 0.8}
\definecolor{dark_violet}{rgb}{0.4, 0.1, 0.4}
\definecolor{light_violet}{rgb}{0.8, 0.1, 0.8}
\definecolor{dark_orange}{rgb}{0.4, 0.4, 0.1}
\definecolor{light_orange}{rgb}{0.8, 0.6, 0.1}

\newcommand{\bm}[1]{\mathbf{#1}}

\begin{document}

\newcommand{\PREPRINTYEAR}{2022}
\newcommand{\PREPRINTPUBLISHER}{IEEE}

\onecolumn
\pagenumbering{gobble}
{
  \topskip0pt
  \vspace*{\fill}
  \centering
  \LARGE{%
    \copyright{} \PREPRINTYEAR~\PREPRINTPUBLISHER\\\vspace{1cm}
	Personal use of this material is permitted.
	Permission from \PREPRINTPUBLISHER~must be obtained for all other uses, in any current or future media, including reprinting or republishing this material for advertising or promotional purposes, creating new collective works, for resale or redistribution to servers or lists, or reuse of any copyrighted component of this work in other works.}
	\vspace*{\fill}
}

\twocolumn 
\pagenumbering{arabic}


\fancypagestyle{empty}{
  \fancyhead{}
  \fancyfoot{}
\renewcommand{\headrulewidth}{0pt}
\fancyhead[LO,RE]{\footnotesize \copyright{} \PREPRINTPUBLISHER, \PREPRINTYEAR. Accepted to ICUAS 2021. DOI: \href{https://doi.org/10.1109/ICUAS54217.2022.9836067}{10.1109/ICUAS54217.2022.9836067}}
\fancyhead[LE,RO]{\footnotesize \thepage}
}

\maketitle
\thispagestyle{empty}
\pagestyle{empty}

\begin{abstract}
Integration of Visual Inertial Odometry (VIO) methods into a modular control system designed for deployment of Unmanned Aerial Vehicles (UAVs) and teams of cooperating UAVs in real-world conditions are presented in this paper.
Reliability analysis and fair performance comparison of several methods integrated into a control pipeline for achieving full autonomy in real conditions is provided.
Although most VIO algorithms achieve excellent localization precision and negligible drift on artificially created datasets,  the aspects of reliability in non-ideal situations, robustness to degraded sensor data, and the effects of external disturbances and feedback control coupling are not well studied.
These imperfections, which are inherently present in cases of real-world deployment of UAVs, negatively affect the ability of the most used VIO approaches to output a sensible pose estimation.
We identify the conditions that are critical for a reliable flight under VIO localization and propose workarounds and compensations for situations in which such conditions cannot be achieved.
The performance of the UAV system with integrated VIO methods is quantitatively analyzed w.r.t. RTK ground truth and the ability to provide reliable pose estimation for the feedback control is demonstrated onboard a UAV that is tracking dynamic trajectories under challenging illumination.  
\end{abstract}

\vspace{1em}

\section*{Multimedia Materials}
\label{sec:multimedia_materials}
The paper is supported by the multimedia materials available at \href{http://mrs.felk.cvut.cz/icuas2022vio}{mrs.felk.cvut.cz/icuas2022vio}.

\section{Introduction}

The recent growth in the availability of Unmanned Aerial Vehicles (UAVs)  increased their applicability in various applications, such as for example infrastructure inspection \cite{Nekovar2021}\cite{Giuseppe2021_ICUAS}, search and rescue \cite{matej_darpa}, and monitoring \cite{Meshcheryakov2022}\cite{Qian2020}\cite{Vivaldini2019}. Current research has been focused on the autonomy of UAVs to allow performing the required task without a human pilot~\cite{Aggarwal2020}.
Real-time UAV localization and state estimation are essential aspects for achieving such autonomous behavior \cite{Jeon2021}.

In most cases of outdoor deployment, the Global Navigation Satellite System (GNSS) is used for localization due to its full availability and easy-to-use approach. However, it limits the UAV deployment to large open areas with direct satellite visibility. Also, its precision is not sufficient for a precise localization required by some applications.
The GNSS precision can be improved with Differential GNSS (DGNSS) or Real-Time Kinematic (RTK), which gives a centimeter precision. Nonetheless, these solutions require specific equipment, for example, a base station for sending GNSS position corrections to the UAV, which further reduces the applicability of such a system.

Another less restricting approach is gathering knowledge of UAV surroundings suitable for localization by onboard sensors, such as LiDAR-based (Hector SLAM \cite{hector} for single-plane LiDARs, and LiDAR Odometry And Mapping - LOAM \cite{loam} for multiple-plane LiDARs), and vision-based methods \cite{Liang2022}\cite{Patel2020}.

\begin{figure}[!t]
    \begin{tikzpicture}
      \node[anchor=south west,inner sep=0] (b) at (0,0) {\adjincludegraphics[width=1.0\linewidth,trim={{0.0\width} {0.01\height} {0.0\width} {0.0\height}}, clip]{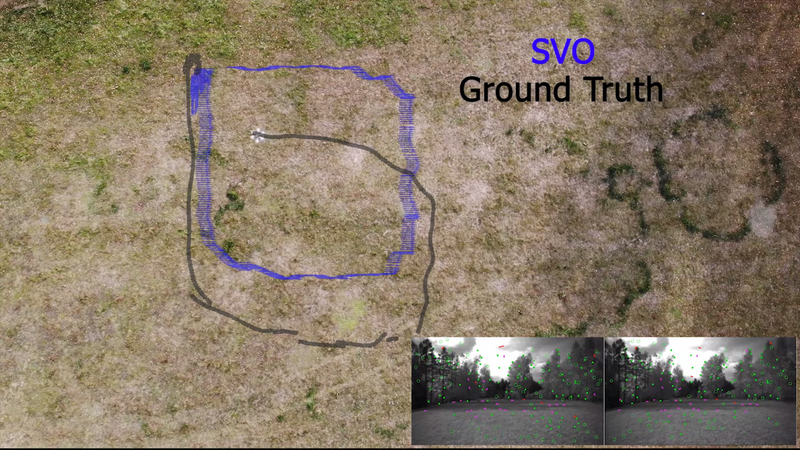}};%
    \begin{scope}[x={(b.south east)},y={(b.north west)}]
      \end{scope}
    \end{tikzpicture}
    \\
    \centering
    \begin{tikzpicture}
      \node[anchor=south west,inner sep=0] (a) at (0,0) {\adjincludegraphics[width=1.0\linewidth,trim={{0.0\width} {0.01\height} {0.0\width} {0.0\height}}, clip]{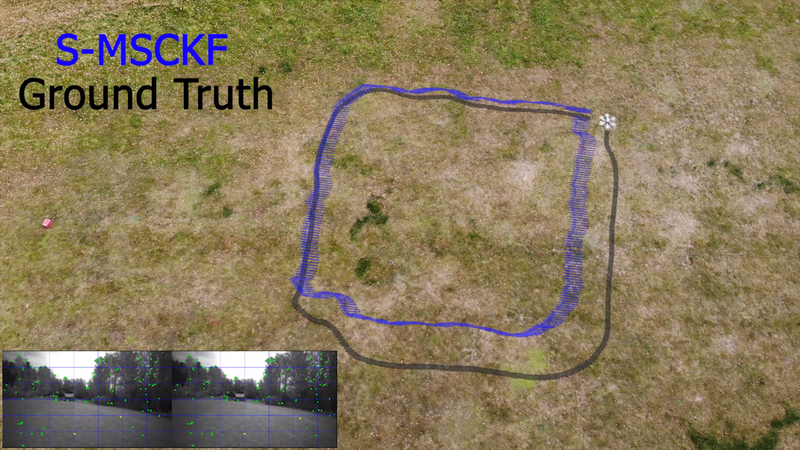}};%
    \begin{scope}[x={(a.south east)},y={(a.north west)}]
      \end{scope}
    \end{tikzpicture}%
    \caption{
      Top-down overview of the experiments from~\autoref{sec:real_feedback}, in which SVO and S-MSCKF algorithms are integrated into the feedback of the control pipeline. 
    \label{fig:rw_exp}
  }
\end{figure}

Although the vision-based approaches are dependent on lighting conditions, using cameras as a localization sensor has several advantages, such as low weight, small size, low cost, and mainly rich data flow. 
Among many possible combinations of sensors, monocular cameras and IMUs (Inertial Measurement Unit) provide the least expensive and lightweight, but still sufficiently robust and precise state estimation, using Visual-Inertial Odometry (VIO) methods. 
VIO algorithms increase the robustness of a Visual Odometry (VO), relying on a camera only, by incorporating information from an IMU to improve motion tracking performance.
VIO approaches exploit the observation that the visual data obtained by the camera can reduce the drift of the IMU data, whereas the IMU data can retrieve the metric scale and roll and pitch angles of the visual data.
Methods that employ stereo cameras instead of monocular cameras estimate the depth of features more accurately, which consequently results in lower metric scale drift.
This increase in performance comes at the cost of a slightly higher computational load.

\subsection{Related works}

Several benchmark studies for visual estimation of UAV state can be found in the literature. 
In \cite{benchmark3}, the benchmark for vision-based odometry is presented and VO methods are compared by evaluating tracking accuracy in terms of the accumulated error (drift) over the full sequence of images, with different resolutions and two lenses with different fields of view.
The dataset does not contain a ground truth and the error is obtained by aligning the beginning and end of trajectory segments from the same part of the environment.

A benchmark of visual-inertial methods regarding computational requirements, translation errors, and absolute translation error on the EuRoC dataset \cite{machine_hall} is proposed by \cite{Delmerico2018}. The authors conclude that the accuracy and robustness of VIO techniques need to be improved.
Other visual odometry comparisons, such as \cite{benchmark2}, are related only to non-inertial methods or non-6DoF variants \cite{benchmark4}.

 KAIST VIO dataset \cite{Jeon2021} serves as a benchmark of Visual-Inertial Odometry methods. The authors show a benchmark test of various visual-inertial odometry algorithms on NVIDIA Jetson platforms showing an analysis of resource usage and RMSE (Root Mean Squared Error) of absolute trajectory error considering image resolution.

In these works, the localization methods are tested on publicly available benchmarks, 
usually based on a custom sensor set. 
We can observe that these studies do not consider reliability aspects in non-ideal real-world situations, robustness to degraded sensor data, and the effects of external disturbances and feedback control coupling.
These imperfections inherently present in any UAV flight in real-world conditions negatively affect the ability of the state-of-the-art VIO approaches to output a sensible pose estimation.

Outlier rejection techniques of the VIO methods that are based on the RPE metric often cannot identify correspondences that cause large orientation errors.
A method reformulating the reprojection process is proposed by \cite{Duan2022} to reject these keypoints.
This method classifies correspondences as inliers and outliers by separating the orientation and translation error components into different dimensions.

Because of the issues raised, this paper identifies critical conditions for a reliable flight under VIO localization and proposes workarounds and compensations for situations in which such conditions cannot be achieved.
Also, we propose a modification of multirotor UAV control and estimation system to integrate the VIO.
For the UAV control, we rely on the well-documented and open-source MRS System \cite{Baca2020} that has been actively used by many researchers in the aerial robotics field. 
This system allowed the experimental verification of methods published in \cite{matej_darpa} \cite{Rocha2021} \cite{control} \cite{matej_surveillance} \cite{plan} that could benefit from the proposed VIO-MRS complex.  


\subsection{Contributions}

The main contributions are as follows:
\begin{itemize}
\item integration of VIO approaches into the UAV control system to enable applications without GNSS access in real-world conditions;
\item  trajectory shaping technique that modifies the dynamics of an input trajectory to reduce motion blur and aggressive tilting motion during which a large part of image features is lost;
\item evaluation of pose estimation precision w.r.t. RTK ground truth under challenging conditions in an outdoor field with distant image features during the dark evening or direct sun;
\item reliability analysis of the VIO methods in the feedback of the SE(3) controller and identification of the failure cases;
\item  optimal placement of the camera considering the feature distribution in the camera image that is often degraded by the high contrast of direct sunlight.
\end{itemize}



\section{Visual-Inertial Odometry Algorithms}

\label{sec:methods}

The state-of-the-art VIO methods can be divided into two principal categories --- filter-based and optimization-based approaches.
The filter-based methods generally rely on a variant of the Extended Kalman Filter (EKF) for estimating the camera pose. 
The IMU measurements of accelerations and angular rates are used for the state propagation of the filter based on a non-linear model, while the update step fuses poses obtained by image feature matching.
An example of filter-based VIO algorithms is S-MSCKF (Stereo Multi-State Constraint Kalman Filter) algorithm \cite{smsckf} and ROVIO (Robust Visual-Inertial Odometry) algorithm \cite{rovio1, rovio2}. 

Optimization-based methods jointly optimize the residuals of image and IMU data measurements to obtain the state estimates. 
There are several optimization-based approaches such as SVO (Semi-Direct Visual Odometry) \cite{Forster2017}, VINS-Fusion (Visual-Inertial System Fusion) \cite{vins_local} or OKVIS (Open Keyframe-Based Visual-Inertial SLAM) \cite{okvis}. 
Judging by the results published in \cite{vins_local}, VINS-Fusion provides a more precise pose estimation than OKVIS.

According to the prior survey, three VIO algorithms have been chosen for the detailed analysis in real-world conditions in this work:
\begin{itemize}
    \item The filter-based S-MSCKF algorithm.
    \item The semi-direct SVO method, which combines feature-based and direct methods with robust probabilistic depth estimation aided by motion priors from the IMU.
   \item The optimization-based VINS-Fusion algorithm, which is a stereo and multi-sensor extension of VINS-Mono.
\end{itemize}

\subsection{S-MSCKF---Stereo multi-state constraint kalman filter}

The S-MSCKF \cite{smsckf} is a stereo variant of the MSCKF \cite{msckf_1} \cite{msckf_2} \cite{msckf_3} approach designed explicitly for UAVs.
The authors show that the S-MSCKF achieves robustness with a modest computational budget for aggressive, three-dimensional maneuvering, and fast flights with speeds reaching up to \SI{17.5}{\meter\per\second}. 
A UAV with S-MSCKF localization can be deployed in both indoor and outdoor environments and also allows indoor/outdoor transitions. 
This method is not suitable for long-distance flights as the uncertainty of the position estimate grows as the UAV travels, which is caused by not observable global position and yaw.
However, the authors demonstrated during an agile flight of over \SI{700}{\meter} that the error is only \SI{3}{\meter}, which is sufficient for most typical applications.

\subsection{SVO---Semi-direct visual odometry}

SVO \cite{svo} is an optimization-based visual odometry algorithm that uses a direct method to track features. 
The method minimizes the photometric error between pixels corresponding to the projected zone of the same 3D point using feature-based methods for joint structure optimization from motion (SfM). 
Furthermore, the robust probabilistic depth estimation algorithm enables tracking pixels on weak corners and edges. 
The authors emphasize the high computation speed of the direct tracking approach, which is possible thanks to the absence of feature extraction and matching steps.
The extension \cite{Forster2017} of the original algorithm includes generalization for multiple-camera systems, motion prior term added to the optimized cost function, and edge features use. These additions further improve the accuracy and especially the robustness to aggressive motion.

\subsection{VINS-Fusion---Visual inertial navigation system}

VINS-Fusion \cite{Qin2019global,Qin2019local} developed as an extension of VINS-Mono \cite{Qin2018}, is based on optimized multi-sensor fusion for theses sensor combinations: stereo camera; stereo camera + IMU; monocular camera + IMU.
The system uses IMU preintegration and feature point observations to achieve precise device self-location. 
In contrast to other state-of-the-art VIO algorithms, VINS-Fusion continuously calibrates the extrinsic parameters between the camera and IMU and allows loop closures to further reduce drift.
An online temporal calibration feature also aims to achieve high accuracy in time offset calibration and system motion estimation~\cite{qin2018online}.
The authors demonstrate that VINS-Fusion achieves locally accurate and globally drift-free pose estimation.
Besides that, the authors indicate that the framework can fuse sensor data with different settings in a unified pose graph optimization and is generalized to use other sensors besides cameras and IMU.


\section{VIO Approaches Designed for UAV Deployment in Demanding Real-world Conditions}
\label{sec:integration}

This part of the paper is focused on aspects required for the deployment of VIO approaches using such low-cost and light-weight sensory setup onboard UAVs in demanding real-world conditions to facilitate transitions of these systems from simulations and laboratories to real UAV missions. We will tackle difficulties of challenging light conditions that may change from direct sun-light to darkness instantaneously due to the transition between indoor and outdoor and obstructions of the sun by objects appearing in the proximity of UAV. The obstacles in the workspace and changing weather conditions may also introduce UAV motion disturbances and vibrations, caused by aerodynamic effects of flying close to objects and wind gusts that are not present in laboratory conditions.

\subsection{Camera orientations}
\label{sec:camera_orientations}

\begin{figure*}[!htb]
    \centering
    \adjincludegraphics[width=0.245\linewidth,trim={{0.0\width} {0.2\height} {0.0\width} {0.0\height}}, clip]{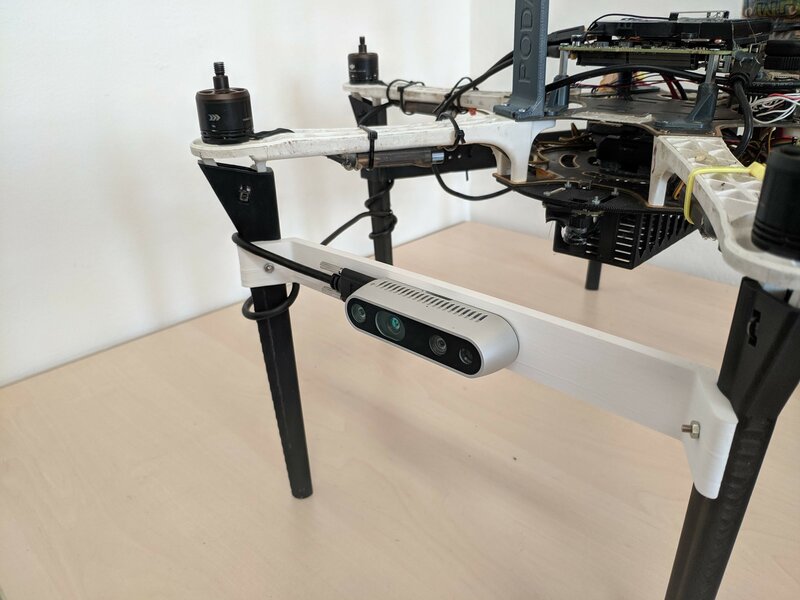}%
    \adjincludegraphics[width=0.245\linewidth,trim={{0.0\width} {0.2\height} {0.0\width} {0.0\height}}, clip]{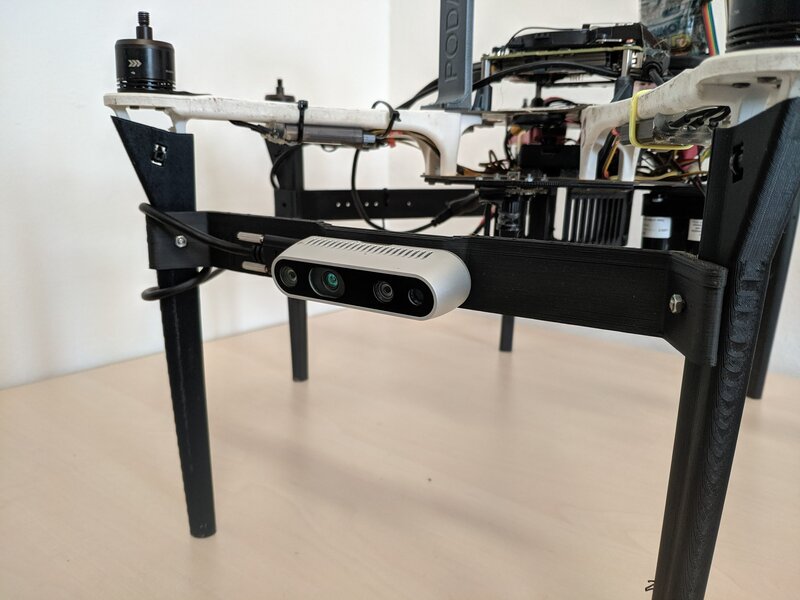}%
    \adjincludegraphics[width=0.245\linewidth,trim={{0.0\width} {0.2\height} {0.0\width} {0.0\height}}, clip]{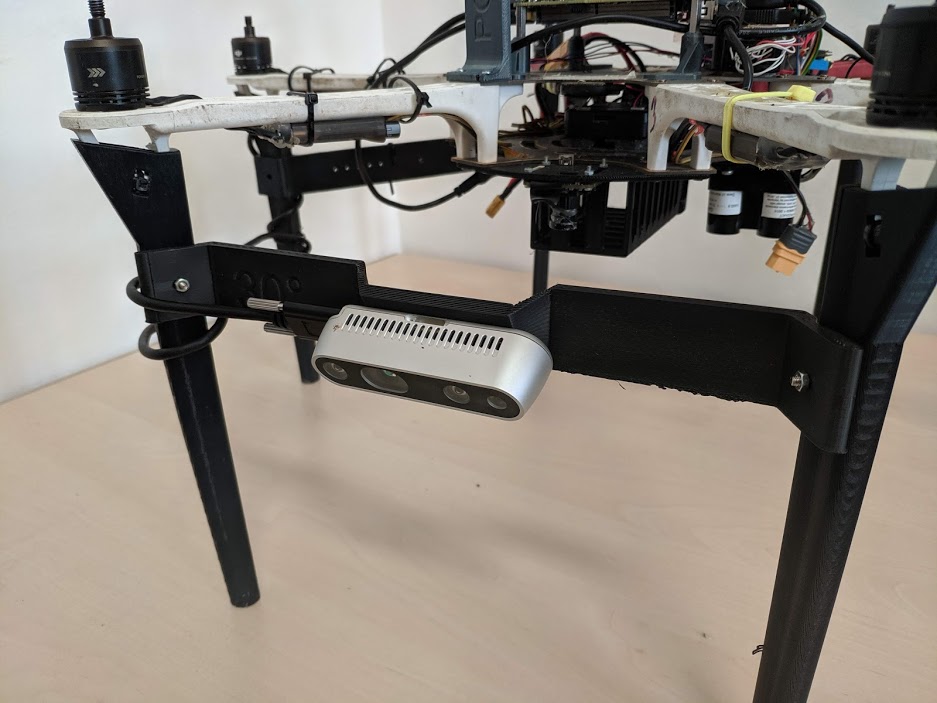}%
    \adjincludegraphics[width=0.245\linewidth,trim={{0.0\width} {0.2\height} {0.0\width} {0.0\height}}, clip]{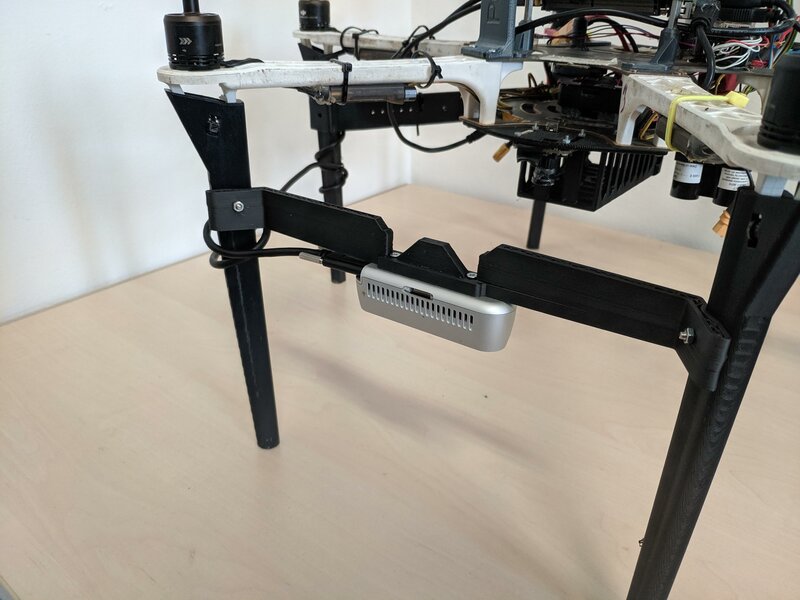}%
    \\
    \adjincludegraphics[width=0.245\linewidth,trim={{0.0\width} {0.035\height} {0.0\width} {0.0\height}}, clip]{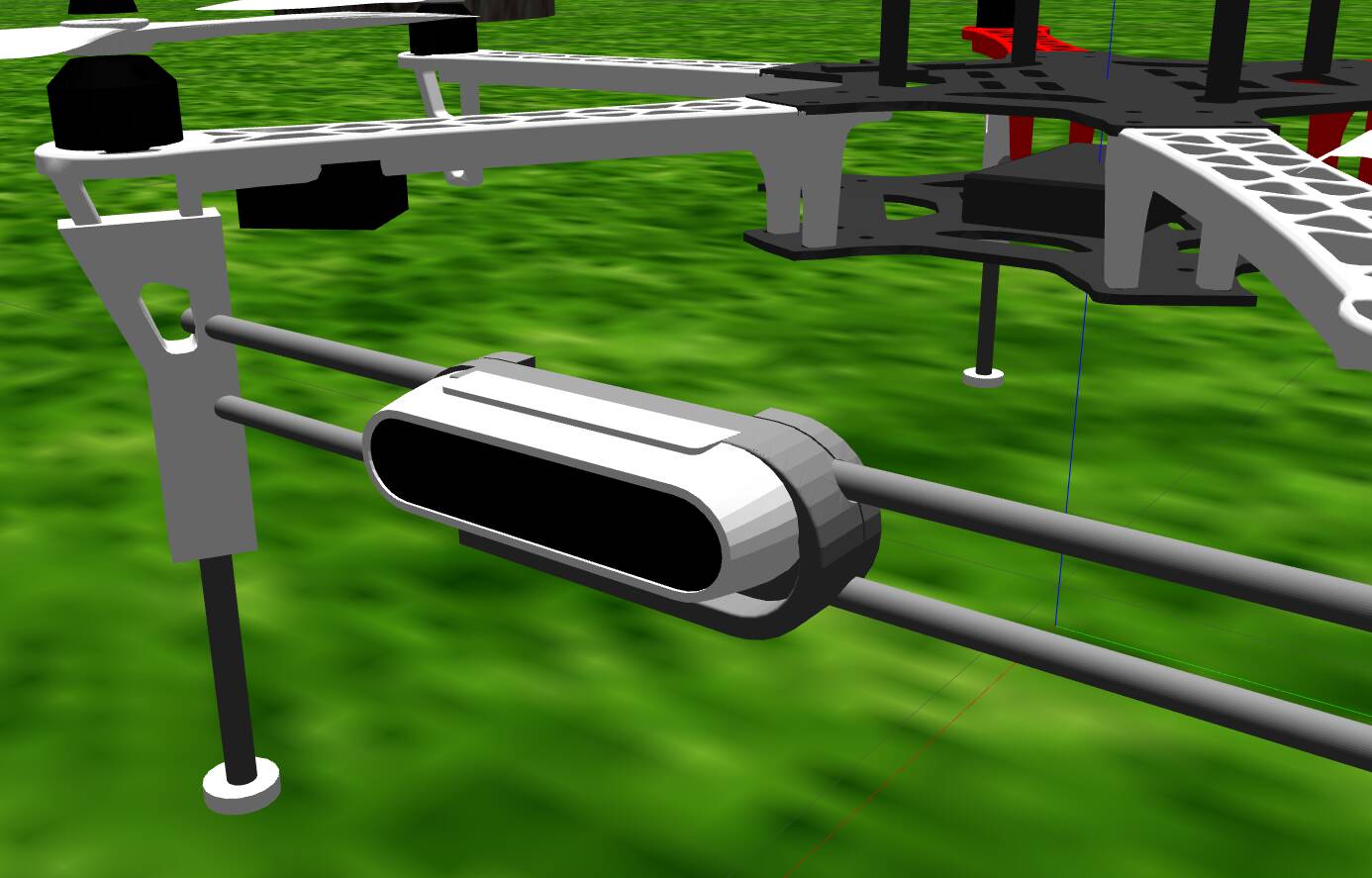}%
    \adjincludegraphics[width=0.245\linewidth,trim={{0.0\width} {0.03\height} {0.0\width} {0.0\height}}, clip]{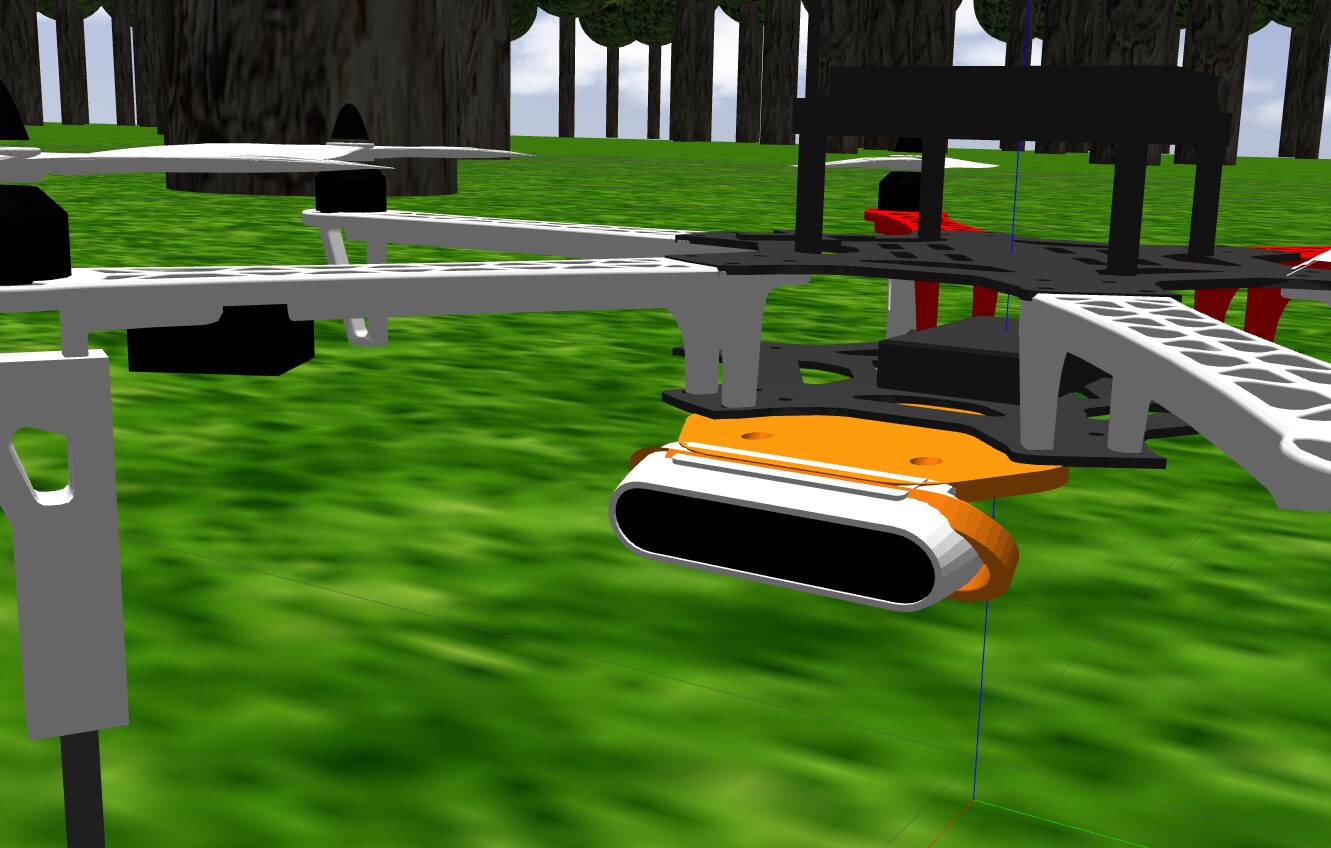}%
    \adjincludegraphics[width=0.245\linewidth,trim={{0.0\width} {0.0\height} {0.005\width} {0.0\height}}, clip]{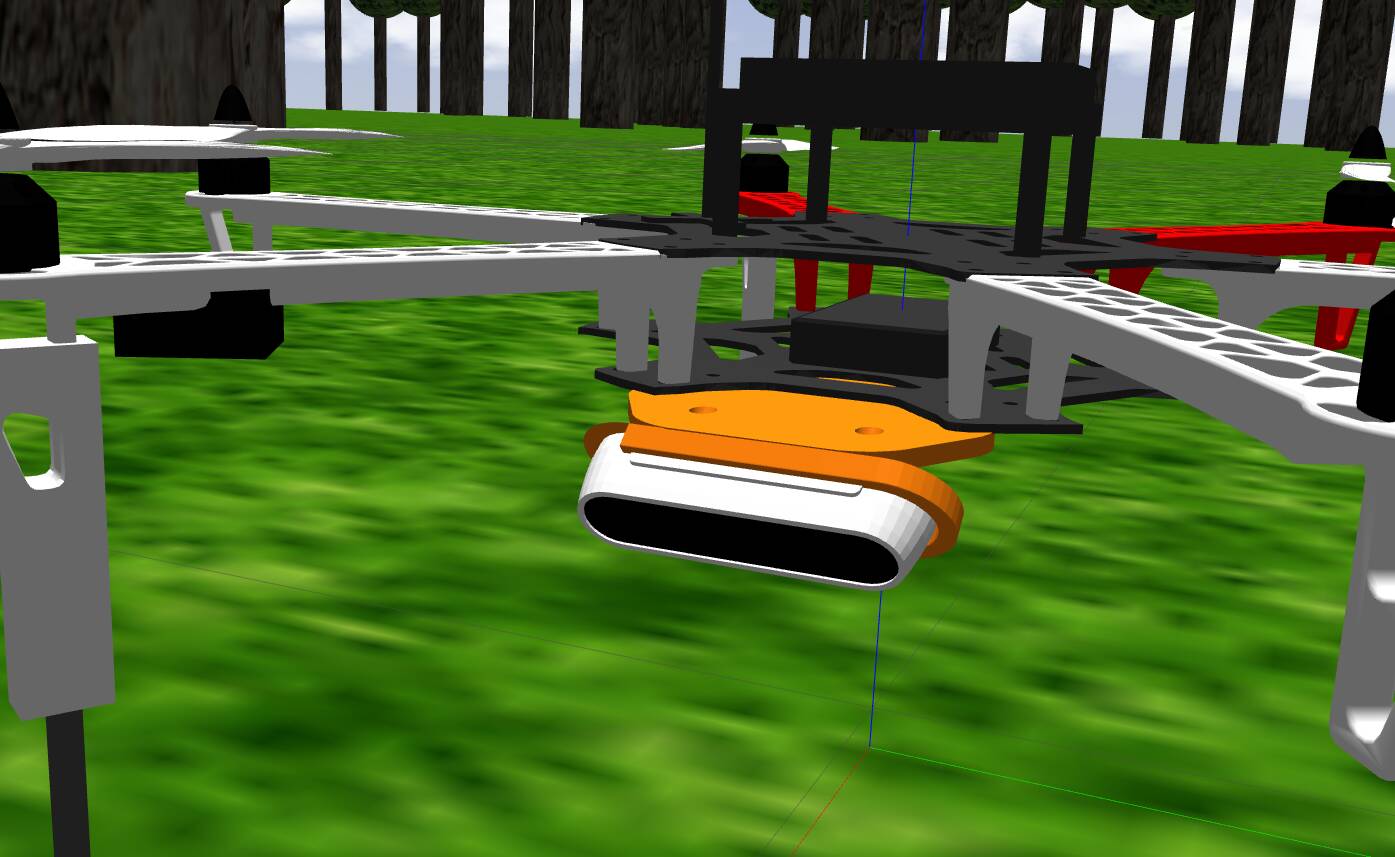}%
    \adjincludegraphics[width=0.245\linewidth,trim={{0.0\width} {0.0\height} {0.055\width} {0.0\height}}, clip]{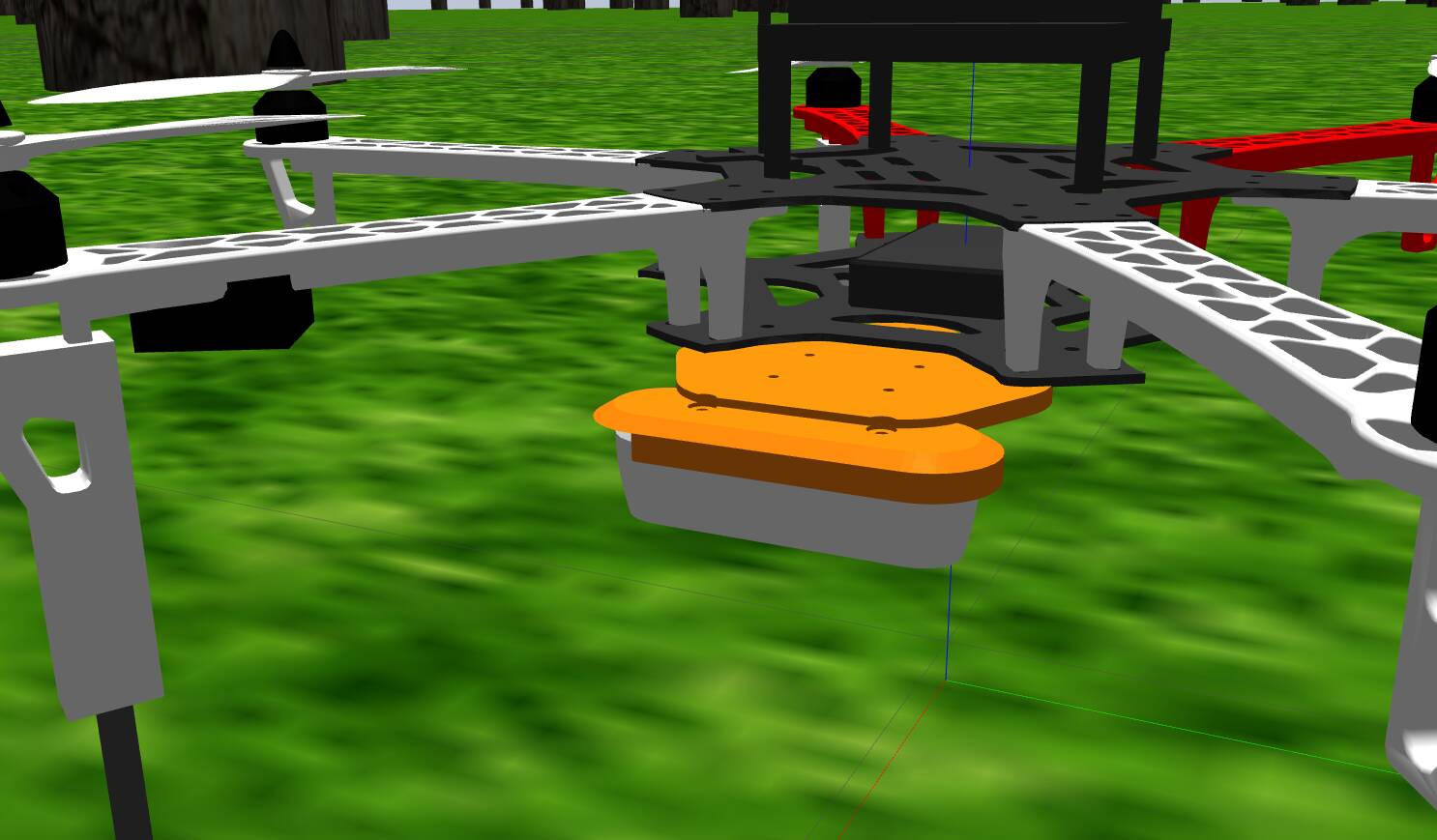}%
    \caption{
      Camera mounting position between the front legs of the real UAV (top row) and the simulated (bottom row) UAV model.
      The orientations of the camera considered in this paper are depicted from left to right in this order: \SI{0}{\degree}, \SI{10}{\degree}, \SI{30}{\degree} and \SI{90}{\degree}.
    \label{fig:camera_attachment}
  }
\end{figure*}

We have theoretically and also experimentally found that the camera orientation influences the number of visual features that can be detected in the image stream during different phases of flight, as well as the ability to track these features under aggressive maneuvers even more in the real environment.
Due to some mechanical constraints of real robots, such as the required placement of the camera between the front legs of the UAV, the pitch angle of the camera changes the part of the scene that is captured.
Four pitch angles are evaluated in this paper: \SI{0}{\degree} (forward), \SI{10}{\degree}, \SI{30}{\degree}, \SI{90}{\degree} (downward).
The position of the camera as well as all the mounting orientations in both real and simulated UAV platforms are shown in~\autoref{fig:camera_attachment}.

The advantage of the forward orientation of the camera is the number of available features during translation motions.
However, this setup might suffer from rapid illumination changes and high contrast, especially in the outdoor environment in sunny weather. 
Besides that, fast rotations in the yaw angle might lead to losing track of many features.

The ratio of ground features with a pitched camera is higher, and their detection is possible even from higher altitudes.
Thus the pitched camera partially avoids the unreliable and unstable features in the distance and in the sky.
More ground features are successfully tracked between frames during fast rotations than with the forward-looking camera.

The down-looking orientation gives a considerable advantage in stable feature detection during challenging lighting conditions. 
Ground features are easy to track even during fast yaw rotations.
However, fast translation and roll/pitch motions at low altitudes are challenging due to rapid scene changes, as many features are lost between frames.
Most importantly, taking off is a challenging task for the down-looking camera due to features being too close to the camera. 
However, if the UAV is taking off from the feature-rich ground, this problem is not significant.

Similar observations led manufacturers of commercial drones to equip their systems with a set of cameras (in some setups up to 12) pointing to all directions to increase reliability in the challenging conditions mentioned above.
The proposed work is aimed at designing a less sensory demanding setup relying on a single stereo camera with IMU for deployment in real environments, which could allow us to reach the size in the order of centimeters of fully autonomous UAVs in near future.


\subsection{Integration into the MRS UAV control system}

\begin{figure}[ht]
  \centering
  \resizebox{1.0\linewidth}{!}{\input{Figures/pipeline_diagram.tex}}
  \caption[Control pipeline diagram of the UAV]{
    Diagram of the control pipeline. 
    The reference trajectory serves as a setpoint for the MPC tracker, which outputs a command for the non-linear $SE(3)$ controller. 
  The non-linear controller produces the orientation and thrust reference for the embedded attitude controller.
    IMU$^1$ symbolizes the variant when the IMU data comes from the same source as images, e.g, the RealSense D435i camera. 
    IMU$^2$ means that the IMU data is from the flight control unit (built-in IMU), e.g., in the Gazebo simulator.
    }
  \label{fig:pipeline}
  \vspace{-1.0em}
\end{figure}
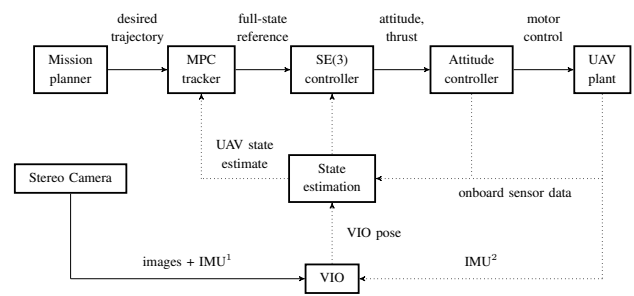

The MRS UAV system, designed by our team for experimental verification of multi-robot approaches, is composed of the control pipeline and state estimation shown in~\autoref{fig:pipeline}.
The \textit{Mission planner} module generates desired trajectories based on the specific mission or task, in this case, the trajectories are square-shaped sequences of positions with sampling corresponding to the desired velocities.
These sequences are processed in the \textit{MPC tracker} \cite{mpc_tracker}, which generates a dynamically feasible full-state reference based on the dynamic model of the UAV.
The \textit{SE(3) controller} employs geometric state feedback to produce thrust and attitude reference for the \textit{Attitude controller} embedded in the Pixhawk autopilot, which in turn controls the speed of the motors propelling the UAV.

The \textit{State estimation} module fuses various sensor measurements in a bank of Kalman filters to obtain hypotheses of lateral position, altitude, and orientation.
After a valid hypothesis is found, the full-state estimate for the \textit{SE(3) controller} feedback is formed.
If no valid hypothesis exists, the control system performs a swift emergency landing, to prevent dangerous motion resulting from diverging state estimate.
This safety feature is critical when testing VIO algorithms in feedback in challenging conditions, in which the algorithms are expected to fail often.

The VIO output is integrated as another sensor entering the \textit{State estimation} module among GNSS, RTK, IMU, barometer, and laser rangefinder.
A Kalman filter with a point-mass model up to the \nth{2} derivative provides such estimate by propagating the accelerations from IMU through the model with position and velocity, if available, corrections from the VIO.
The orientation must be supplied by the IMU because the delay introduced by the VIO computation, image capture, and transfer would destabilize the delay-sensitive attitude controller.
The control system requires a state estimate at a specific rate (MRS UAV system works at \SI{100}{\hertz}), while VIO algorithms provide measurements at the rate of the camera or IMU.
During the update step of the filter, the model state can be propagated to a requested time at desired rate regardless of the VIO rate.

\subsection{Trajectory shaping}
\label{sec:traj_shapping}

\begin{figure}[htp]
 \centering
  \begin{tikzpicture}
      \node[anchor=south west,inner sep=0] (a) at (0,0){\adjincludegraphics[width=0.49\linewidth,trim={{0.15\width} {0.00\height} {0.20\width} {0.115\height}}, clip]{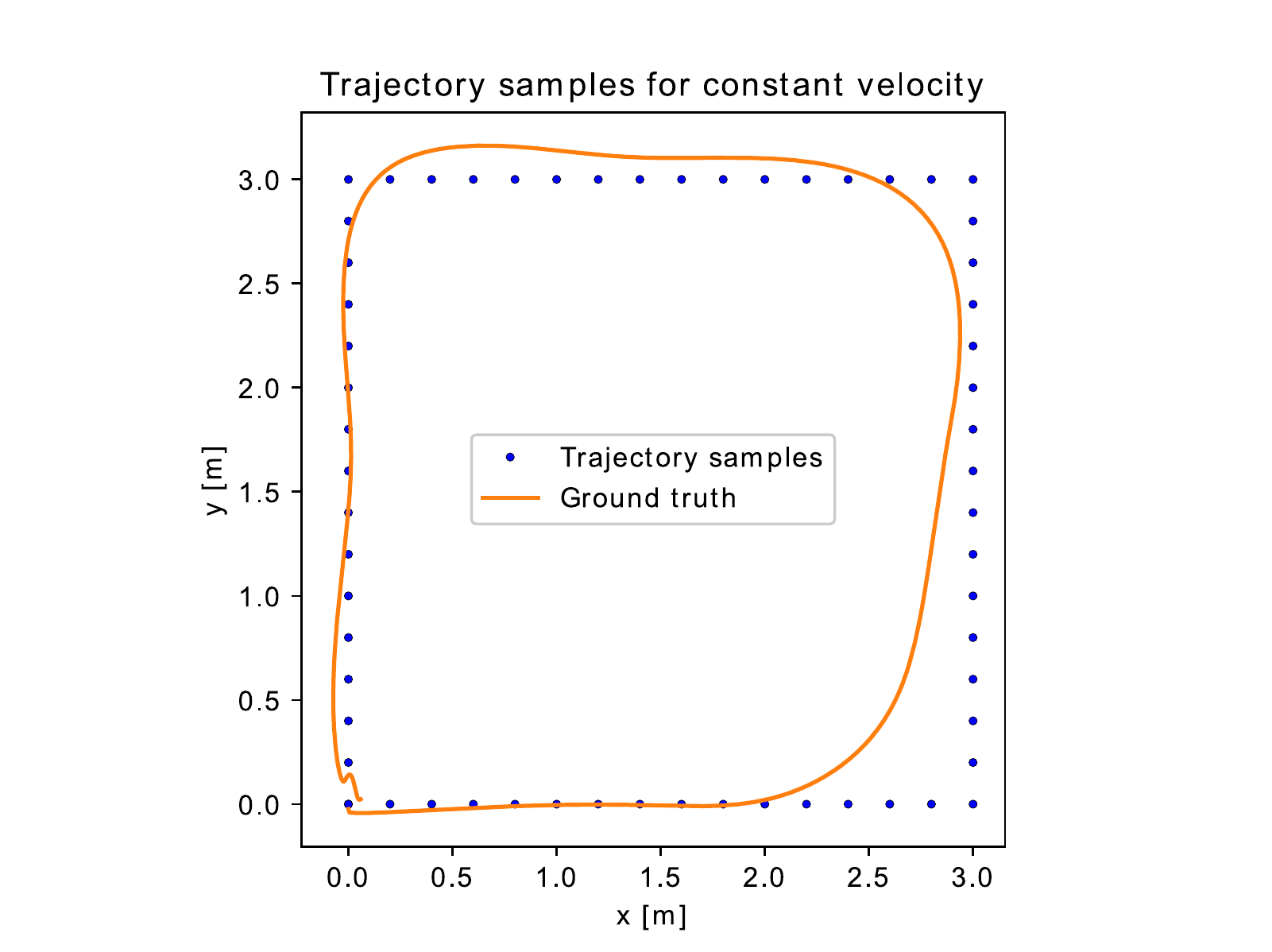}};
    \begin{scope}[x={(a.south east)},y={(a.north west)}]
      \node[fill=black, fill opacity=0.0, text=black, text opacity=1.0] at (0.3, 0.3) {\textbf{(a)}};
    \end{scope}
      \end{tikzpicture}
  \begin{tikzpicture}
    \node[anchor=south west,inner sep=0] (b) at (0,0){\adjincludegraphics[width=0.49\linewidth,trim={{0.15\width} {0.00\height} {0.20\width} {0.115\height}}, clip]{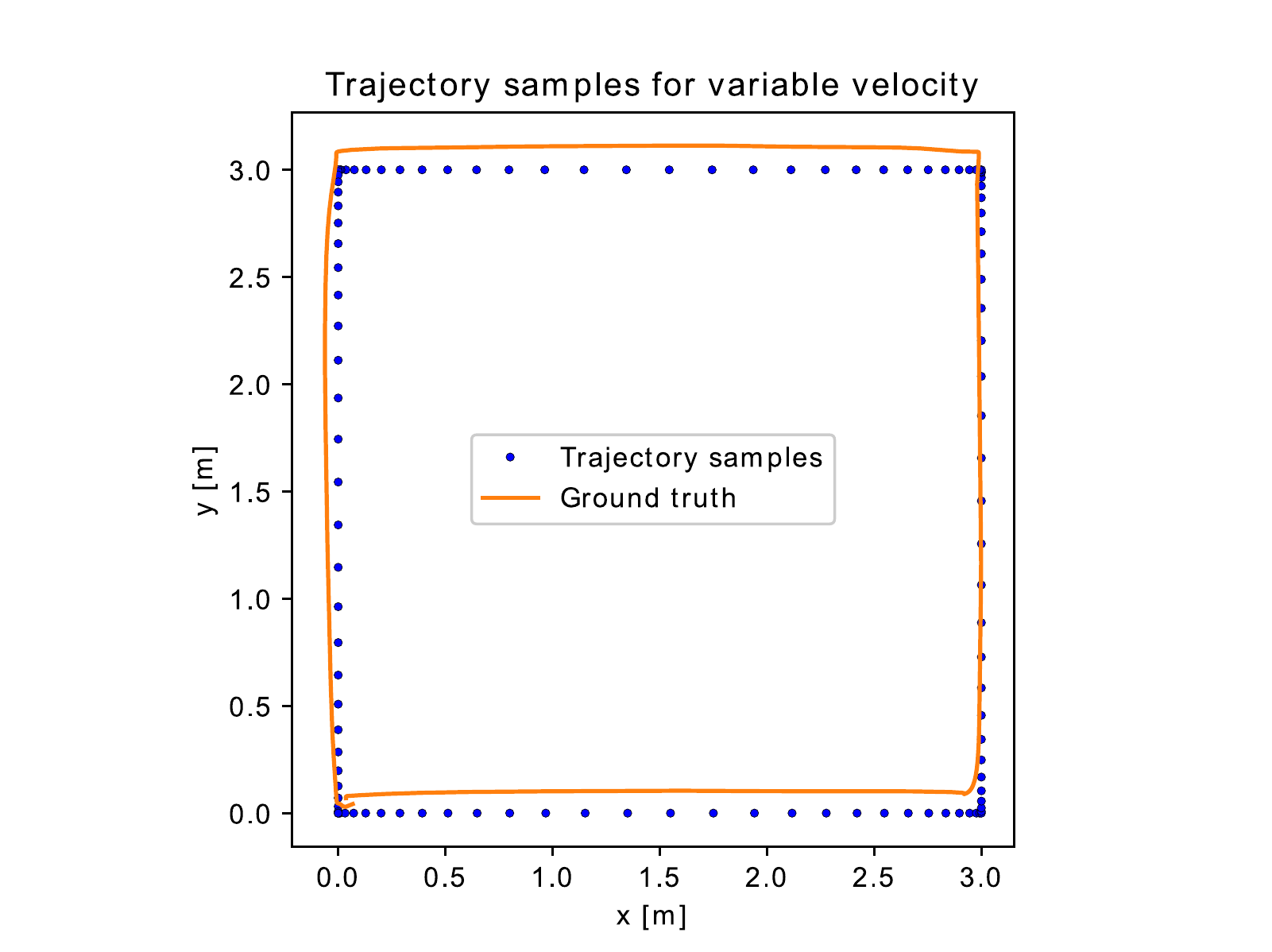}};
    \begin{scope}[x={(a.south east)},y={(a.north west)}]
      \node[fill=black, fill opacity=0.0, text=black, text opacity=1.0] at (0.3, 0.3) {\textbf{(b)}};
    \end{scope}
      \end{tikzpicture}
  \caption{
    Trajectory with constant velocity profile (a) violates acceleration constraints during the UAV flight. 
    The UAV platform used in this flight is described in~\autoref{sec:uav_platform}.
    The smoothed trajectory (b) is constrained to \SI{0.4}{\meter\per\second\squared}, which improves tracking precision, and prevents aggressive tilting that would result in losing image features.
  } 
 \label{fig:orig_trajectory}
\end{figure}

A typical approach to generating trajectories involves the equidistant sampling of a geometric path to obtain a trajectory with a constant velocity profile.
These are valid trajectories that are accepted by the MPC tracker, which transforms them into a reference that respects the dynamic constraints of the UAV, but the reference might deviate from the original trajectory.
Moreover, the maximum acceleration of the UAV is so high that it induces aggressive tilts of the UAV, which negatively impacts the accuracy of VIO due to the disappearance of a large part of tracked features from the Field of view (FoV) of the camera.
\autoref{fig:orig_trajectory}a shows how the UAV tracks a simple square-shaped trajectory with a constant velocity profile.

The proposed trajectory shaping method includes constraints of VIO approaches into the UAV motion planning and adapts the trajectory according to the UAV dynamics so that the motion is respecting the requirements of reliable usage of the VIO-based control mechanism in real-world conditions.
First, critical points where the UAV motion (mainly acceleration) exceeds motion constraints required by a particular VIO approach are found in the trajectory.
Then, an iterative approach of adding trajectory samples in the proximity of these points is applied and the motion constraints are consequently evaluated and verified using the state estimation and motion prediction approaches in \cite{mpc_tracker}.
The smoothed trajectory that respects the constraints of VIO approaches and the tracking of this trajectory by the UAV are shown in~\autoref{fig:orig_trajectory}b. 
Details on the VIO motion constraints specification and determination can be found in the following sections.

\section{Meaningful Evaluation of Vision-based Approaches in Demanding Real-world Conditions}
\label{sec:Practical}

A crucial aspect of comparing localization methods, which is one of the key contributions of this paper, is evaluating the translation and rotation drifts, i.e., the deviation of estimated position from ground truth, over a period of time. 
This section defines the metrics used to evaluate the selected state-of-the-art VIO methods w.r.t. precise ground truth, describes the hardware and software of the experimental platform that was used to carry out the experiments in real-world conditions, and introduces the specific scenarios that were used for the comparison of the algorithms to provide fair and meaningful (w.r.t. real deployment) evaluation.

\subsection{Ground truth}
\label{sec:ground_truth}

\begin{figure}[!htb]
    \centering
    \begin{tikzpicture}
      \node[anchor=south west,inner sep=0] (a) at (0,0) {\adjincludegraphics[width=0.5\linewidth,trim={{0.0\width} {0.0\height} {0.0\width} {0.0\height}}, clip]{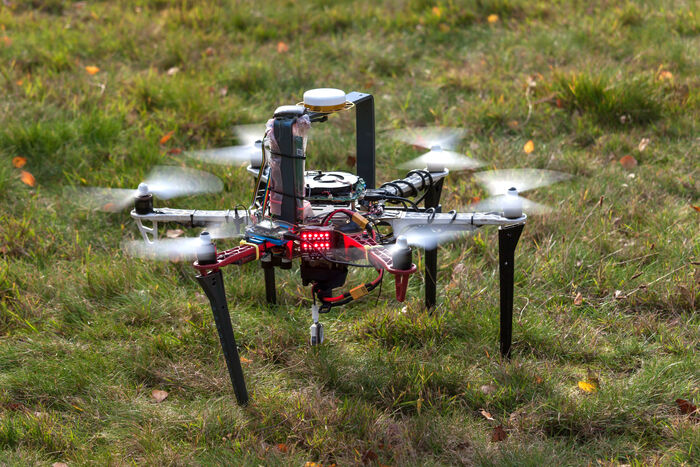}};%
    \begin{scope}[x={(a.south east)},y={(a.north west)}]
      \node[fill=black, fill opacity=0.0, text=white, text opacity=1.0] at (0.10, 0.15) {\textbf{(a)}};
      \end{scope}
    \end{tikzpicture}%
    \begin{tikzpicture}
      \node[anchor=south west,inner sep=0] (b) at (0,0) {\adjincludegraphics[width=0.5\linewidth,trim={{0.0\width} {0.0\height} {0.0\width} {0.0\height}}, clip]{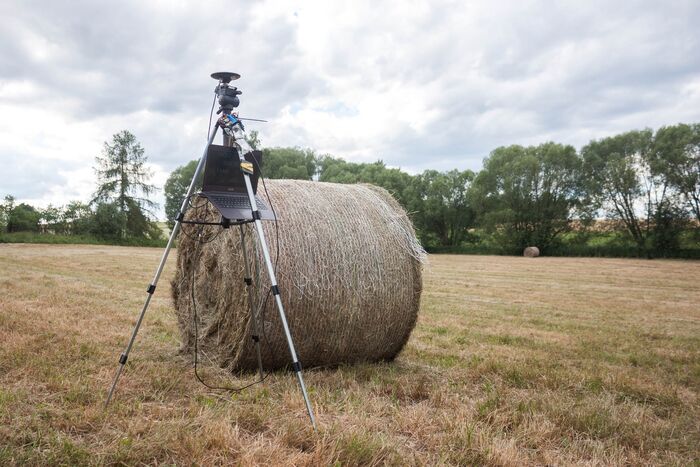}};%
    \begin{scope}[x={(b.south east)},y={(b.north west)}]
      \node[fill=black, fill opacity=0.0, text=white, text opacity=1.0] at (0.10, 0.15) {\textbf{(b)}};
      \end{scope}
    \end{tikzpicture}
    \caption{
      The RTK receiver on top of the UAV (a) and the RTK base station (b) sending corrections for the UAV.
    \label{fig:rtk}
  }
\end{figure}

The ground truth for all experiments with the UAV platform was obtained using an RTK GNSS receiver mounted on the UAV (\autoref{fig:rtk}a) in combination with a fixed ground base station (\autoref{fig:rtk}b) that was sending corrections to the receiver.
Thanks to these corrections and phase shift analysis, the RTK system achieves horizontal precision of up to \SI{1}{\centi\meter} in the most precise mode of operation\,---\,RTK FIX.
However, when the receiver and the base station do not see enough common satellites, the mode of operation can degrade into a worse precision RTK FLOAT solution with a standard deviation of up to \SI{0.5}{\meter}.
Vertical positioning has two times larger error than horizontal measurements, and as described in~\autoref{sec:uav_platform} the UAV has other sensors that accurately measure its altitude.
Thus, the vertical component of the ground truth is not used for the evaluation of the algorithms.
The accuracy of orientation estimated by the VIO algorithms is also not considered explicitly in the evaluation as the orientation error implicitly causes translation error during movement \cite{ate}.  
Moreover, to fully measure the orientation of the UAV, at least 3 RTK receivers would have been mounted on the UAV, which was not feasible with the platform used to conduct the experiments.

\subsection{Evaluation metrics}
\label{sec:Evaluation_Metrics}

The experiments conducted to evaluate the performance of the VIO algorithms focus on the evaluation of the lateral error, i.e., the $x$ and $y$ components of the UAV position in the reference frame defined by the pose of the UAV at the start of the experiment.
Two metrics that are commonly used for evaluating localization methods\,---\,the ATE (absolute trajectory error) and RPE (relative pose error)\,---\,as defined in~\cite{ate} were used to compare the VIO algorithms during real deployment.

The VIO odometry and the ground truth trajectories in time are defined as a series of poses $\bm{P}_1,...,\bm{P}_n \in SE(3)$ and $\bm{Q}_1,...,\bm{Q}_n \in SE(3)$, respectively.
As the VIO and RTK positions are usually in the different inertial systems, the Horn method \cite{horn} was adopted to align the trajectories prior to the evaluation. 
The transformation $\bm{S}$ is used to align the VIO trajectory $\bm{P}_i$ to the ground truth trajectory $\bm{Q}_i$ giving absolute trajectory error $\bm{F}_i$ at time step $i$:

\begin{equation}
  \bm{F}_i := \bm{Q}_i^{-1}\bm{S}\bm{P}_i.
  \label{eq:trajectory_error}
\end{equation}

\subsubsection{ATE---Absolute trajectory error}
The ATE metric is used as the primary evaluation metric, comparing the whole estimated trajectory with the reference ground truth.
The metric is given by the RMSE (Root Mean Square Error) calculated over the previously aligned trajectory:
\begin{equation} \label{eq:rmse_ate}
  \text{ATE}(\bm{F}_{1:n}) := \left( \frac{1}{n} \sum_{i=1}^{n} \left\lVert\text{trans}(\bm{F}_i)\right\rVert^{2}\right)^{\frac{1}{2}},
\end{equation}
where $\text{trans}(\bm{F}_i)$ symbolizes the translation component of the trajectory error $\bm{F}_i$.


\subsubsection{RPE---Relative pose error}
The VIO algorithms often have a local drift in the set of IMU and camera measurements.
This type of error is reflected in the RPE metric, which is calculated similarly to ATE over the previously aligned trajectory.
Contrary to ATE, RPE captures the local drift by evaluating the error over a fixed time interval $\Delta$.
The local pose error is calculated as:

\begin{equation}
  \bm{E}_i := \big( \bm{Q}_i^{-1}\bm{Q}_{i+\Delta}\big)^{-1} \big(\bm{P}_i^{-1}\bm{P}_{i+\Delta} \big).
 \label{eq:error_RPE}
\end{equation}

A total of $m = n - \Delta$ local pose errors is obtained from $n$ pairs of $\bm{P}_i$, $\bm{Q}_i$ poses.
Then, RMSE is calculated over all $m$ errors $\bm{E}_i$:

\begin{equation} \label{eq:rmse_rpe}
  \text{RPE}(\bm{E}_{1:n},\Delta) := \left( \frac{1}{m} \sum_{i=1}^{m} \left\lVert\text{trans}(\bm{E}_i)\right\rVert^{2}\right)^{\frac{1}{2}}.
\end{equation}

The value of the interval $\Delta$ is set to \SI{1}{\second} in all experiments presented in this paper.


\subsection{UAV platform}
 \label{sec:uav_platform}
 
\begin{figure}[ht] 
  \centering
  \includegraphics[width=1.0\linewidth]{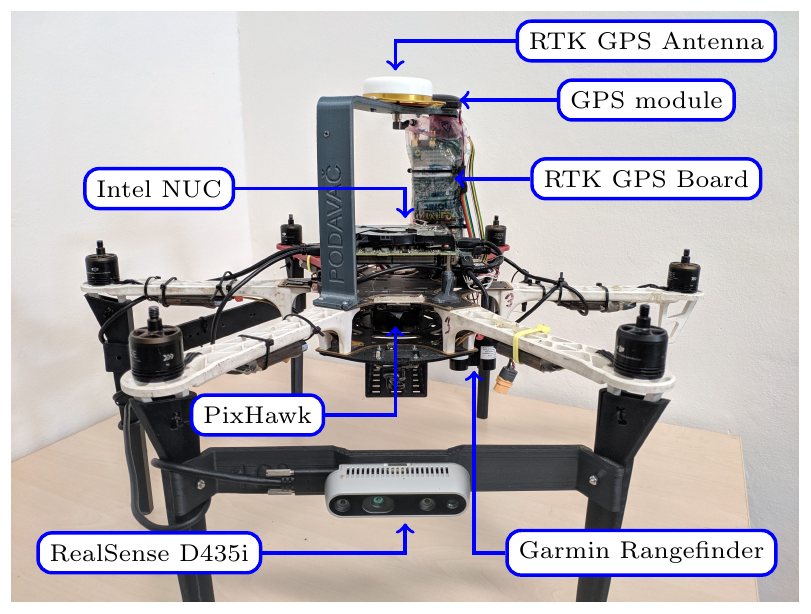}
  \caption[The UAV platform]{
    The UAV platform that was used for the experimental evaluation.
  }
  \label{fig:dron}
\end{figure}

The UAV platform (\autoref{fig:dron}) that was used for the experiments is based on a DJI F550 hexacopter frame equipped with E310 DJI motors.
PixHawk 4 flight control unit handles the low-level control of attitude and attitude rate of the UAV with the help of measurements from the integrated IMU. 
In addition, Pixhawk contains a barometer that is fused with accelerations from the IMU and range measurements from the downward-facing laser rangefinder to estimate the altitude of the UAV.

On top of the UAV is mounted the RTK GNSS receiver, which in tandem with the stationary base station provides position estimates with \SI{1}{\centi\meter} precision.
These position estimates are used as the ground truth as described in~\autoref{sec:ground_truth}.

The continuous support along with easy setup, small form factor, a wide range of possible operating modes, and mainly meeting all requirements of all considered VIO approaches were the reasons for choosing the Intel RealSense D435i camera for the experimental evaluation of the VIO algorithms.
RealSense D435i is a compound camera consisting of rolling shutter RGB 1920$\times$1080@\SI{30}{\hertz} sensor, stereo pair of global shutter greyscale 1280$\times$720@\SI{30}{\hertz} sensors, and Bosch BMI055 IMU unit.
In the presented experiments, only the stereo camera and IMU in RealSense were used and the system can be operated with any other camera satisfying the requirements of a particular VIO method.

All onboard software runs on the Intel NUC 8i7BEH computer that is capable of running 8 threads at \SI{2.7}{\giga\hertz} base clock frequency.


\subsection{Gazebo simulator}
\label{sec:gazebo_simulator_setup}

The UAV platform relying on the above-mentioned MRS system, along with the whole sensor suite, may be simulated in the realistic Gazebo simulator \cite{gazebo} including the PX4 flight control stack to facilitate the safe integration of new methods into the control pipeline.
The PX4 model allows parametrizing the IMU parameters, most importantly the noise density and random walk values of the measured angular velocities and linear accelerations.
As such, the simulated IMU is a realistic model of the IMU of the Pixhawk flight controller onboard the UAV platform.
The RTK does not need to be simulated, as the simulator outputs the exact position of the UAV model in the world, which is used as ground truth instead of the RTK.

The camera plugin allows simulation of the RealSense D435i camera with added noise to emulate the challenging visual conditions. 
However, the images from the simulated camera are not degraded by insufficient illumination and direct sunlight, which we are interested to evaluate the most.
Also, the motion blur, autoexposure, and nonlinearities of the lens are not modeled.
Although the simulator is a quick and cheap way to reject parameter combinations that provide mediocre results even under favorable conditions, real-world experiments are necessary to provide meaningful results.


\subsection{Testing scenarios}
\label{sec:testing_scenarios}

The considered VIO algorithms were first compared in simulations with various combinations of camera orientations, frame rates, resolutions, and flight heights to find the optimal parameters for further experiments with the real UAV platform.
Then the algorithms were evaluated onboard a UAV using the ATE and RPE metrics during demanding experiments, where the UAV was flying a square trajectory in an outdoor environment with challenging lighting conditions and distant features.
Such trajectory enabled repeatability and imposed sharp turns as well as straight segments that both can be problematic using different VIO methods. 
The last set of experiments that were conducted both in simulation and onboard with the real UAV, evaluates the reliability of the algorithms when used in the feedback of the control loop of the UAV as better results of metrics~\autoref{sec:Evaluation_Metrics} obtained by a passive data gathering do not necessarily imply better feedback control performance.

A trajectory of a square shape with a total length of \SI{80}{\meter} was generated for the evaluation.
The velocity of UAV flight influences the distance that the image features move between individual camera frames.
To analyze the influence of the velocity on the reliability of the algorithms, the experiments are conducted on trajectories generated with \SI{1}{\meter\per\second}, \SI{2}{\meter\per\second} and \SI{5}{\meter\per\second} velocities.

\section{Experiments}
\label{sec:experiments}

This section first presents simulation experiments aimed at finding optimal camera parameters.
Then, selected VIO algorithms are compared using data gathered by onboard sensors of real UAVs.
Lastly, the performance of the VIO approach in control feedback is analyzed in simulations and real outdoor environments.

\subsection{Frame rate optimization}

\label{sec:frame_rate}
In this test, different FPS rates are tested for \SI{90}{\degree} camera orientation to evaluate the feature tracking performance during rapid movement.
Generally, the higher the frame rate, the higher the computational time.
However, the higher frame rate reduces the distance of feature movement between frames, which could improve the performance during fast movements.
Thus, the desired UAV velocity is \SI{5}{\meter\per\second} in this experiment to highlight the impact of increased frame rate.
The camera resolution is set to 640$\times$360 and the UAV altitude is kept at \SI{3}{\meter}.
The results of this experiment are summarized in~\autoref{tab:high_fps_test}.

\begin{table}[!htb] 
  \scriptsize
  \centering
  \setlength{\tabcolsep}{3pt}
\renewcommand{\arraystretch}{1.0}
\begin{tabular}{cccc|cc|cc} \toprule
  \multirow{3}{0.9cm}{\centering \textbf{Camera FPS}} & \multirow{3}{1.0cm}{\centering \textbf{UAV velocity [\SI{}{\meter\per\second}]}} & \multicolumn{2}{c}{\textbf{S-MSCKF}} & \multicolumn{2}{c}{\textbf{SVO}} & \multicolumn{2}{c}{\textbf{VINS-Fusion}} \\
  \cline{3-8}
  \noalign{\smallskip}
  \noalign{\smallskip}
& & ATE & RPE & ATE & RPE & ATE & RPE \\ \midrule
  30 & 5 & 1.4281$_2$ & 0.1974$_2$ & 2.2822 & 0.3562 & 9.2172$_1$   & 0.6869$_1$\\ \midrule
  60 & 5 & \textbf{0.3141}$_1$ & \textbf{0.1534}$_1$ & 0.4384 & 0.1048 & \textbf{0.0567}   & \textbf{0.0712}\\ \midrule
  90 & 5 & 1.0513$_1$ & 0.9427$_1$ & \textbf{0.4095} & \textbf{0.0912} & x & x\\ \bottomrule
\end{tabular}
\caption{
  Results from the frame rate evaluation experiment in the simulation environment.
  Each scenario has been repeated 5 times with the same parameters to verify the robustness and precision.
  The value in the table is the median from these five trials.
  The subscript, if present, indicates the count of successful test repetitions giving the calculated values.
  If not present, all repetitions were successful.
  The x symbol means none of the tests were successful.
  The best frame rate for each algorithm is in bold.
}
\label{tab:high_fps_test}
\end{table}

The S-MSCKF algorithm struggles with fast feature changes at \SI{30}{\hertz}.
Increasing the frame rate to \SI{60}{\hertz} improves both metrics, but at \SI{90}{\hertz} the error starts to rise again as the algorithm approaches the limit of the available CPU resources.

The SVO results proved its authors' claim that the higher camera frame rate reduces the computational cost per frame.
This allows operations with low errors even for scenarios with \SI{5}{\meter\per\second} UAV velocity.
Both \SI{60}{\hertz} and \SI{90}{\hertz} rates improved the precision of the SVO result, but the increase from \SI{30}{\hertz} to \SI{60}{\hertz} had a much higher impact (ATE decreased more than 5 times) than the increase from \SI{60}{\hertz} to \SI{90}{\hertz} (ATE decreased by \SI{7}{\percent}).

Similarly to S-MSCKF, VINS-Fusion performed best when the frame rate was increased from \SI{30}{\hertz} to \SI{60}{\hertz} but did not converge at \SI{90}{\hertz}.


\subsection{Evaluation using data gained onboard of real UAVs outdoor}
\label{sec:real}

In this experiment, the VIO algorithms were compared on the data captured by onboard sensors during tracking of trajectories with \SI{0.5}{\meter\per\second}, \SI{1}{\meter\per\second} and \SI{2}{\meter\per\second} desired velocities.
During these experiments, direct sunlight significantly influences the camera.
The comparison is repeated for all four camera orientations from~\autoref{sec:camera_orientations} to see if higher pitch angles improve the performance by not pointing into the sun. 
Furthermore, IMU measurements are severely degraded by the vibrations induced by UAV propellers during flight.
The accelerometer noise parameters of VINS-Fusion and S-MSCKF had to be increased, otherwise, their estimates started diverging.

\begin{table}
  \centering

  \scriptsize
\begin{tabular}{ccccc} \toprule
  \multirow{3}{1.3cm}{\centering \textbf{Camera orientation}} & \multirow{3}{1.3cm}{\centering \textbf{UAV velocity} [\SI{}{\meter\per\second}]} & \multirow{3}{*}{\textbf{S-MSCKF}} & \multirow{3}{*}{\textbf{SVO}} & \multirow{3}{*}{\textbf{VINS-Fusion}} \\\\
  \\ \midrule
    \SI{0}{\degree}  & 0.5 & \textbf{0.5533} & 1.6090 & 1.7885 \\
    \SI{0}{\degree}  & 1   & \textbf{0.6878} & 2.0226 & 1.8431 \\
    \SI{0}{\degree}  & 2   & \textbf{0.3553 }& 2.5472 & 0.6605 \\ \midrule
    \SI{10}{\degree} & 0.5 & \textbf{0.2797} & 2.7256 & 0.4704 \\
    \SI{10}{\degree} & 1   & \textbf{0.6952} & 2.4049 & 0.9081 \\ \midrule
    \SI{30}{\degree} & 0.5 & x      & 5.5443 & \textbf{0.5744} \\
    \SI{30}{\degree} & 1   & x      & 4.8295 & \textbf{1.2863} \\ \midrule
    \SI{90}{\degree} & 0.5 & \textbf{0.7793} & 9.1213 & 3.0201$_1$ \\
    \SI{90}{\degree} & 1   & \textbf{0.4793} & 6.0972 & 5.3223$_1$ \\ \bottomrule
\end{tabular}
\caption{
  ATE results (in meters) of camera orientation evaluation experiment using three different datasets from real UAV flights.
  The value in the table is the median from these 3 trials.
  The subscript, if present, indicates the count of successful test repetitions giving the calculated values.
  If not present, all repetitions were successful.
  The x symbolizes that none of the repetitions were successful.
  The best algorithm with each camera orientation and UAV velocity is in bold.
}
\label{tab:real_test}
\end{table}

\autoref{tab:real_test} indicates the results of algorithms on datasets taken by the real UAV.
S-MSCKF algorithm clearly outperformed the other two algorithms in all tests, except in the \SI{30}{\degree} camera orientation.

VINS-Fusion, the best candidate from the simulation experiments, worked well in all trials except the \SI{90}{\degree} camera orientation. 
The proximity of all visible features during takeoff seems to misalign the VINS-Fusion scale at startup, giving worse ATE results.

The SVO underperformed in every test case, which seems to be due to the poor quality of the camera images. 
Even though enough features are detected, the scale of the trajectory is not correct and the loosely-coupled IMU configuration cannot improve the estimation performance.

Surprisingly, S-MSCKF failed to converge with the \SI{30}{\degree} pitch angle of the camera, since the D435i camera suffered from occasional flickering, especially under outdoor sunlight.
This was caused by an issue in the auto-exposure controller of the camera, which has been fixed since the dataset creation.
Although it was a hardware problem of the camera, it shows that S-MSCKF is more sensitive to fast exposure changes than the other tested algorithms.

In general, all camera orientations worked well on the real UAV datasets. 
The advantage of the \SI{10}{\degree} pitch of the camera is the absence of direct sunlight on the camera while still keeping most of the features in front of the UAV.
This improvement is especially prominent in the results of the VINS-Fusion.

\subsection{Feedback in simulation}
\label{sec:sim_feedback}

In this test, the VIO algorithms are tested in the feedback of the UAV control system in simulation. 
The results for \SI{0}{\degree}, \SI{10}{\degree} and \SI{30}{\degree} pitch angles of the camera were not vastly different in prior simulation tests due to the absence of sudden exposure changes and other visual issues that are present in the real-world dataset from~\autoref{sec:real}.
That is why only \SI{0}{\degree} and \SI{90}{\degree} orientations are tested in the feedback. 

SVO has proved its robustness in a higher frame rate, so \SI{60}{\hertz} rate is used. 
S-MSCKF and VINS-Fusion algorithms did not perform well with high frame rates in~\autoref{sec:frame_rate}, hence they are evaluated at \SI{30}{\hertz} camera rate.

The altitude is set to \SI{5}{\meter} for the \SI{90}{\degree} camera orientation and \SI{3}{\meter} for the \SI{0}{\degree} camera orientation, which should achieve the best performance based on initial simulations.
All results are summarized in~\autoref{tab:simulation_feedback}. 

\begin{table}[!htb]
\scriptsize
  \centering
  \setlength{\tabcolsep}{3pt}
\begin{tabular}{cccc|cc|cc} \toprule
  \multirow{3}{1.2cm}{\centering \textbf{Camera orientation}} & \multirow{3}{1.0cm}{\centering \textbf{UAV velocity} [\SI{}{\meter\per\second}]} & \multicolumn{2}{c}{\textbf{S-MSCKF}} & \multicolumn{2}{c}{\textbf{SVO}} & \multicolumn{2}{c}{\textbf{VINS-Fusion}} \\
  \cline{3-8}
  \noalign{\smallskip}
  \noalign{\smallskip}
 & & ATE & RPE & ATE & RPE & ATE & RPE \\ \midrule
    \SI{0}{\degree}   & 1  & 0.6473 & 0.0871 & 0.6182 & 0.0610 & \textbf{0.3369} & \textbf{0.0451} \\
    \SI{0}{\degree}   & 2  & \textbf{0.3202} & 0.0967 & 0.7480 & 0.0781 & 0.4145 & \textbf{0.0944} \\
    \SI{0}{\degree}   & 5  & 0.4518 & 0.1390 & 1.2714 & 0.0994 & \textbf{0.3961} & \textbf{0.0773} \\ \midrule
    \SI{90}{\degree}  & 1  & 0.0647 & 0.0504 & 0.3470 & 0.1041 & \textbf{0.0402} & \textbf{0.0294} \\
    \SI{90}{\degree}  & 2  & 0.1185 & 0.0630 & 0.2481 & 0.0973 & \textbf{0.0378} & \textbf{0.0371} \\
    \SI{90}{\degree}  & 5  & 0.2437 & 0.1426 & 0.3894 & 0.1131 & \textbf{0.1078} & \textbf{0.0631} \\ \bottomrule
\end{tabular}
\caption{The results of feedback control test in the simulation environment.
  Each scenario has been repeated 3 times to avoid outliers.
  The value in the table is the median from these 3 trials.
  The best scenario accomplished for all trials for each algorithm and every camera orientation is in bold.
 }
\label{tab:simulation_feedback}
\end{table}


In general, the best precision was achieved with \SI{90}{\degree} camera orientation and lower velocities. 
But, the UAV could not take off with the VIO control feedback due to the lack of features while staying on the ground. 
So the UAV has to take off using an additional odometry sources, such as GNSS, to initialize the VIO.
The \SI{0}{\degree} camera orientation results are slightly worse than the \SI{90}{\degree} camera orientation due to the faster disappearance of image features during fast translation and yaw movement.

VINS-Fusion achieved an almost driftless trajectory with \SI{90}{\degree} camera orientation, which makes it the top candidate for the real deployment.
The stability is assured by using measurements from both visual features and IMU in the optimization process.
S-MSCKF results show that it also performs well with the \SI{90}{\degree} camera orientation.
The stability of the UAV is satisfactory among all velocities for \SI{90}{\degree} camera orientation.
The EKF-based approach of features and IMU fusion improves the stability during rapid movement changes, such as during accelerations when features are changing faster.
SVO precision for \SI{90}{\degree} camera orientation is better than the results for \SI{0}{\degree} camera orientation. 
Contrary to the other two algorithms, the SVO relies primarily on the features.
The IMU measurements are used only for motion corrections, improving the result. 
The number of features decreases rapidly during aggressive maneuvers, resulting in positional drift, which is partially compensated by the higher frame rate.


\subsection{Feedback control with the real UAV} 
\label{sec:real_feedback}

After verifying the feasibility of integrating the algorithms into the feedback control in the simulated scenario, the algorithms were tested on a real UAV.
All tests are performed at the \SI{30}{\hertz} frame rate and the default altitude set to \SI{3}{\meter}.
Except for the \SI{90}{\degree} camera orientation with the \SI{1}{\meter\per\second} UAV velocity, where the altitude is set to \SI{5}{\meter} to facilitate feature tracking during maneuvers involving aggressive tilting.

S-MSCKF algorithm presented satisfying results in the experiment in~\autoref{sec:real} and thus it was possible to test it in the feedback loop of the UAV control system.
SVO algorithm was not as successful, but it was stable enough to test its performance.
Unfortunately, VINS-Fusion could not be safely tested in the feedback loop, because it was unstable.
Hence only S-MSCKF and partly SVO results are presented.
Top-down snapshots of the experiments are shown in~\autoref{fig:rw_exp}.

\begin{figure}
  \centering
    \begin{tikzpicture}
      \node[anchor=south west,inner sep=0] (a) at (0,0) {\adjincludegraphics[width=0.5\linewidth,trim={{0.00\width} {0.00\height} {0.00\width} {0.07\height}}, clip]{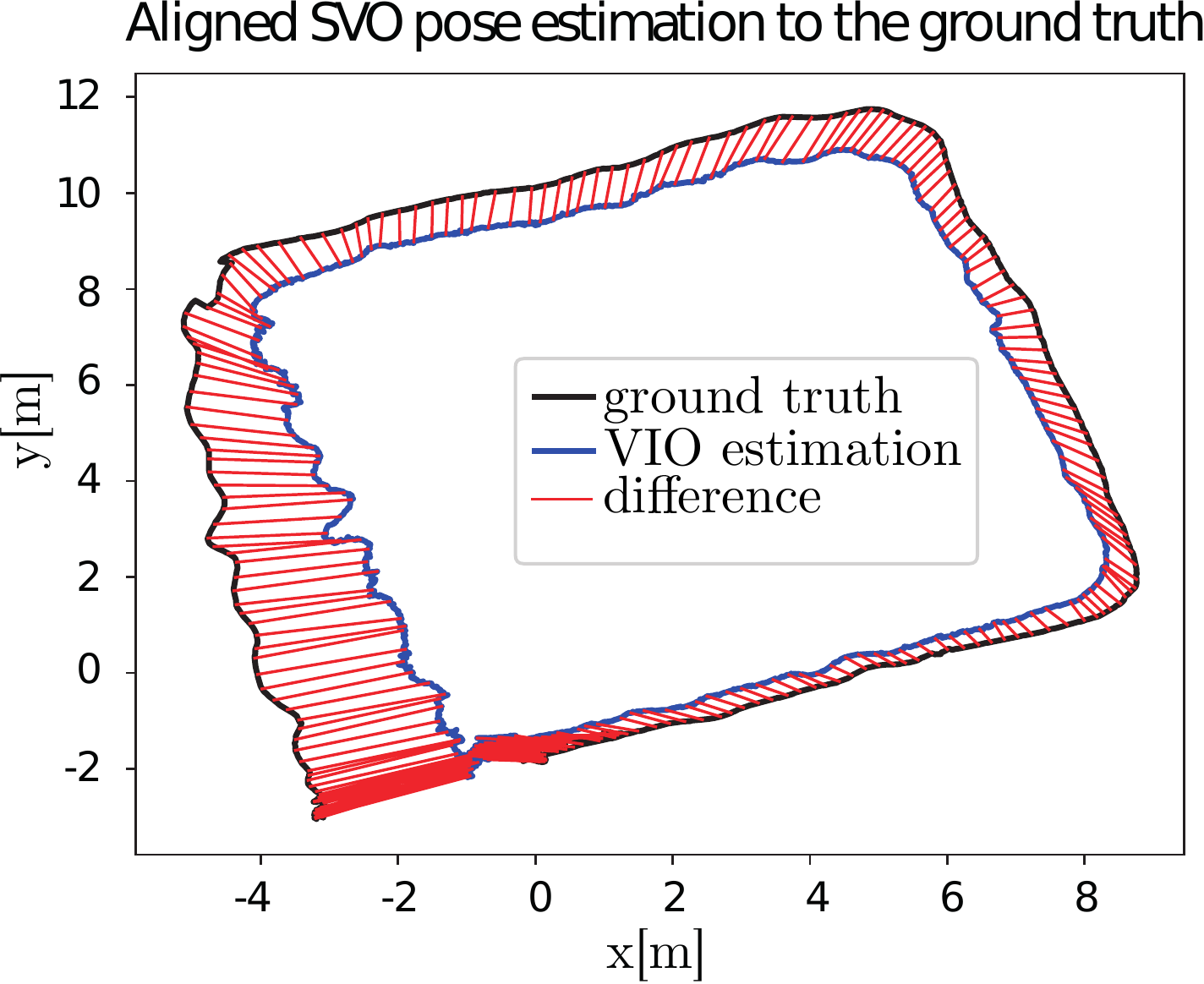}};
    \begin{scope}[x={(a.south east)},y={(a.north west)}]
      \node[fill=black, fill opacity=0.0, text=black, text opacity=1.0] at (0.92, 0.23) {\textbf{(a)}};
      \end{scope}
    \end{tikzpicture}%
    \begin{tikzpicture}
      \node[anchor=south west,inner sep=0] (b) at (0,0) {\adjincludegraphics[width=0.5\linewidth,trim={{0.00\width} {0.00\height} {0.00\width} {0.07\height}}, clip]{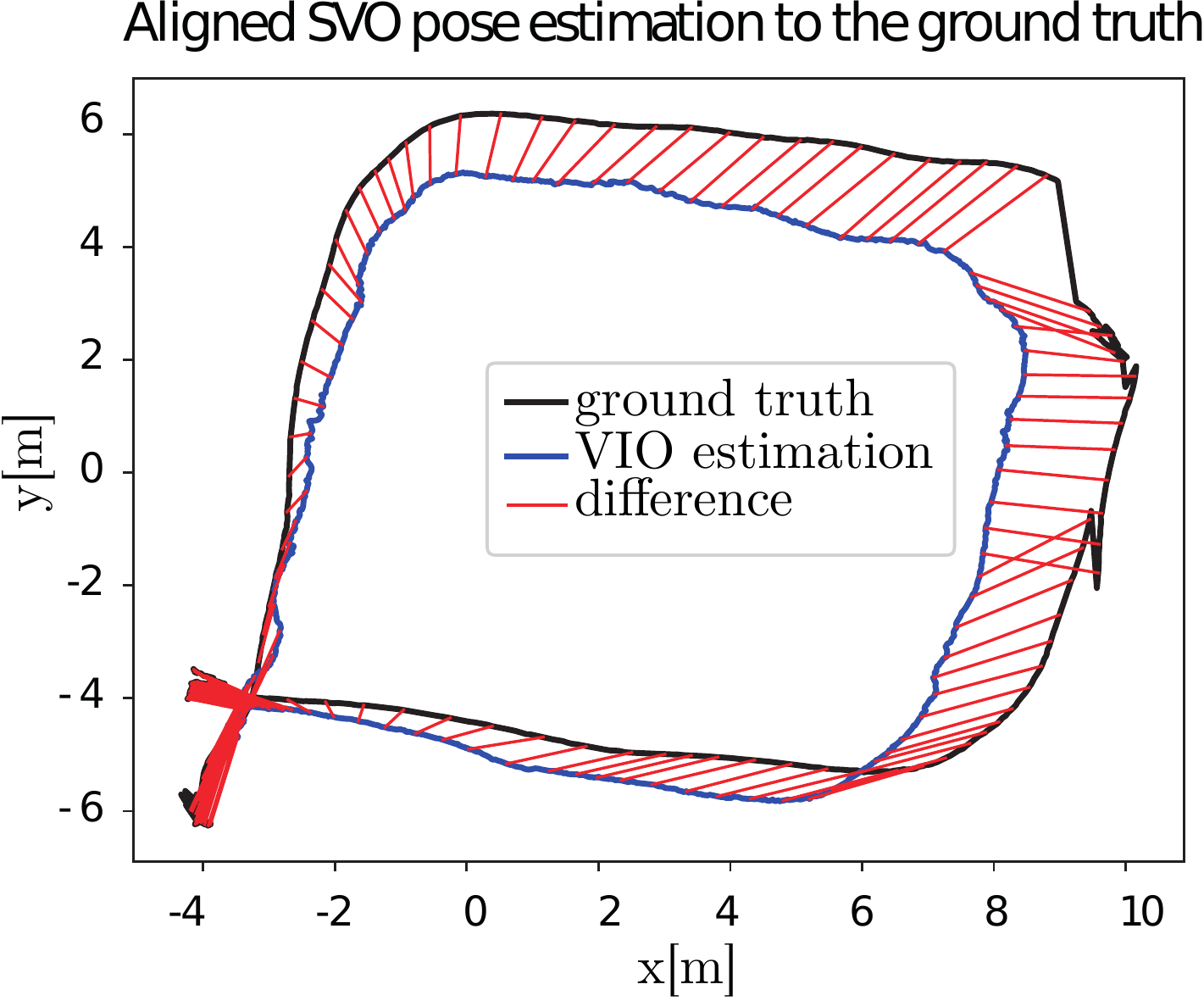}};
    \begin{scope}[x={(a.south east)},y={(a.north west)}]
      \node[fill=black, fill opacity=0.0, text=black, text opacity=1.0] at (0.92, 0.23) {\textbf{(b)}};
      \end{scope}
    \end{tikzpicture}
     \caption[Feedback loop control with SVO algorithm on the real UAV]{
       Trajectories flown with SVO in the feedback of the control system with \SI{0}{\degree} camera orientation.
        ATE of \SI{1.3395}{\meter} was achieved on trajectory (a) with \SI{0.5}{\meter\per\second} desired velocity. 
        Faster trajectory (b) with \SI{1}{\meter\per\second} desired velocity reached \SI{1.3038}{\meter} ATE.
  }
  \label{fig:svo_real_feedback1}
\end{figure}

\begin{figure}
  \centering
    \begin{tikzpicture}
      \node[anchor=south west,inner sep=0] (a) at (0,0) {\adjincludegraphics[width=0.5\linewidth,trim={{0.00\width} {0.00\height} {0.00\width} {0.00\height}}, clip]{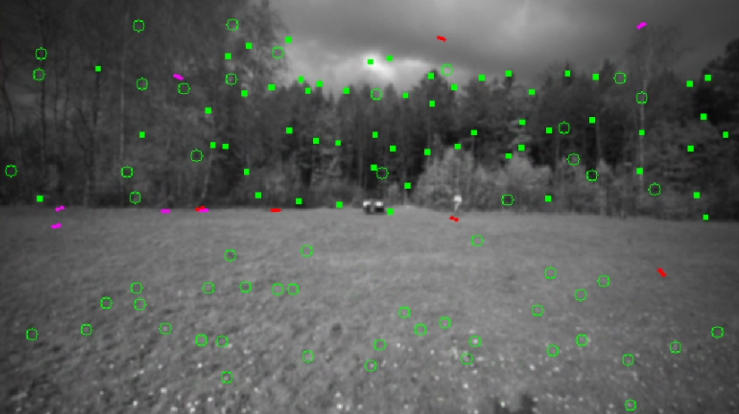}};
    \begin{scope}[x={(a.south east)},y={(a.north west)}]
      \node[fill=white, fill opacity=0.0, text=white, text opacity=1.0] at (0.92, 0.13) {\textbf{(a)}};
      \end{scope}
    \end{tikzpicture}%
    \begin{tikzpicture}
      \node[anchor=south west,inner sep=0] (b) at (0,0) {\adjincludegraphics[width=0.5\linewidth,trim={{0.00\width} {0.00\height} {0.00\width} {0.01\height}}, clip]{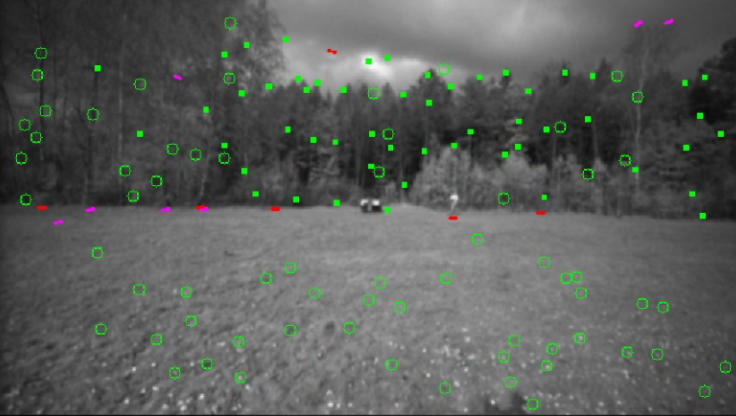}};
    \begin{scope}[x={(a.south east)},y={(a.north west)}]
      \node[fill=white, fill opacity=0.0, text=white, text opacity=1.0] at (0.92, 0.13) {\textbf{(b)}};
      \end{scope}
    \end{tikzpicture}
     \caption[Feedback loop control with SVO algorithm on the real UAV]{
       Left (a) and right (b) camera images with SVO features during the feedback experiment with \SI{0}{\degree} camera orientation.
  }
  \label{fig:svo_real_feedback_images}
\vspace{-1.0em}
\end{figure}

Despite the volatile performance during the real dataset experiments, SVO pose estimation was usable in the feedback loop, as shown in~\autoref{fig:svo_real_feedback1}. 
The camera images with the features detected by the algorithm can be seen in~\autoref{fig:svo_real_feedback_images}.
However, the position error was huge due to inaccurate estimation of the metric scale.

\begin{figure}
  \centering
    \begin{tikzpicture}
      \node[anchor=south west,inner sep=0] (a) at (0,0) {\adjincludegraphics[width=0.495\linewidth,trim={{0.00\width} {0.00\height} {0.00\width} {0.00\height}}, clip]{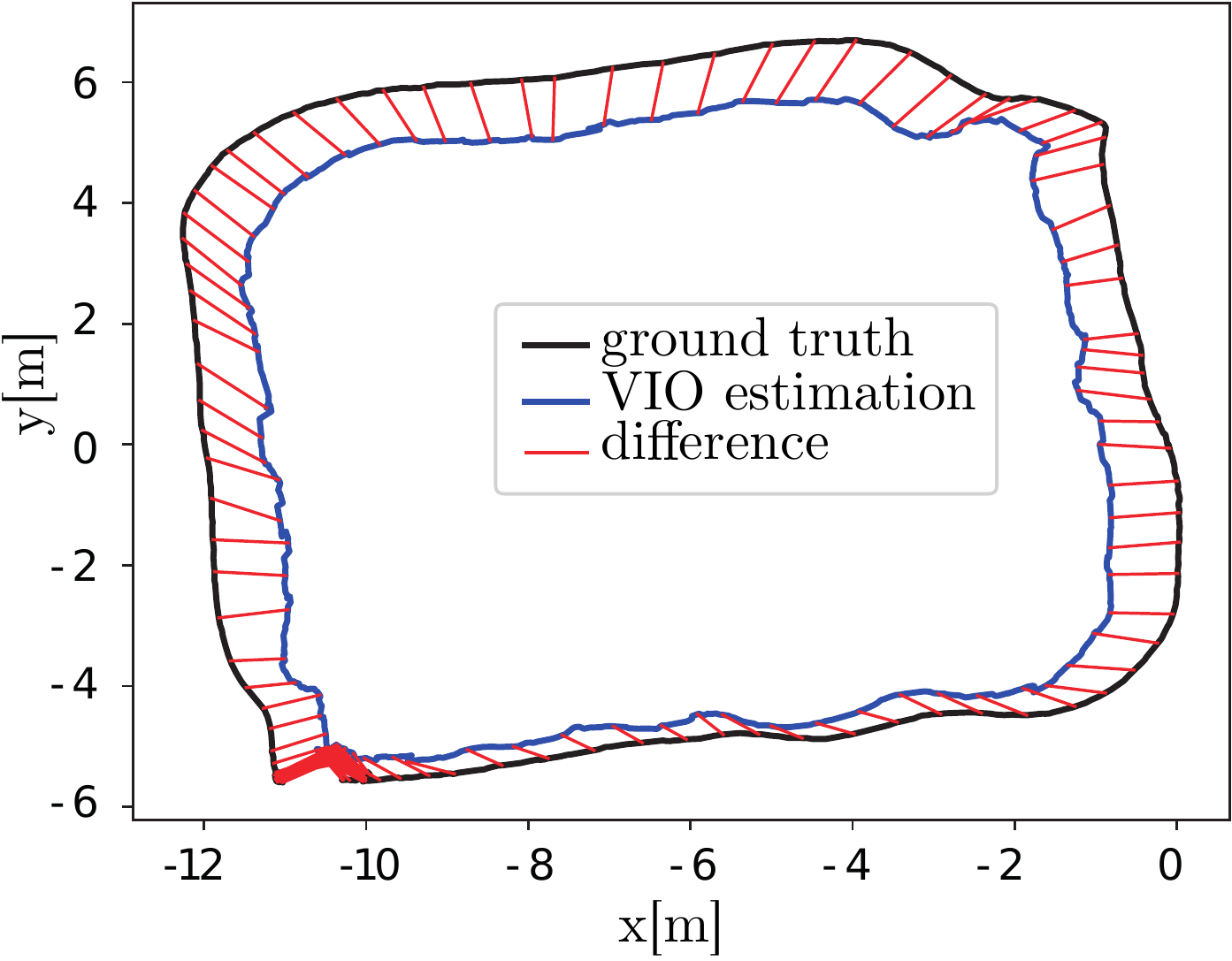}};
    \begin{scope}[x={(a.south east)},y={(a.north west)}]
      \node[fill=black, fill opacity=0.0, text=black, text opacity=1.0] at (0.93, 0.22) {\textbf{(a)}};
      \end{scope}
    \end{tikzpicture}%
    \begin{tikzpicture}
      \node[anchor=south west,inner sep=0] (b) at (0,0) {\adjincludegraphics[width=0.495\linewidth,trim={{0.00\width} {0.00\height} {0.00\width} {0.00\height}}, clip]{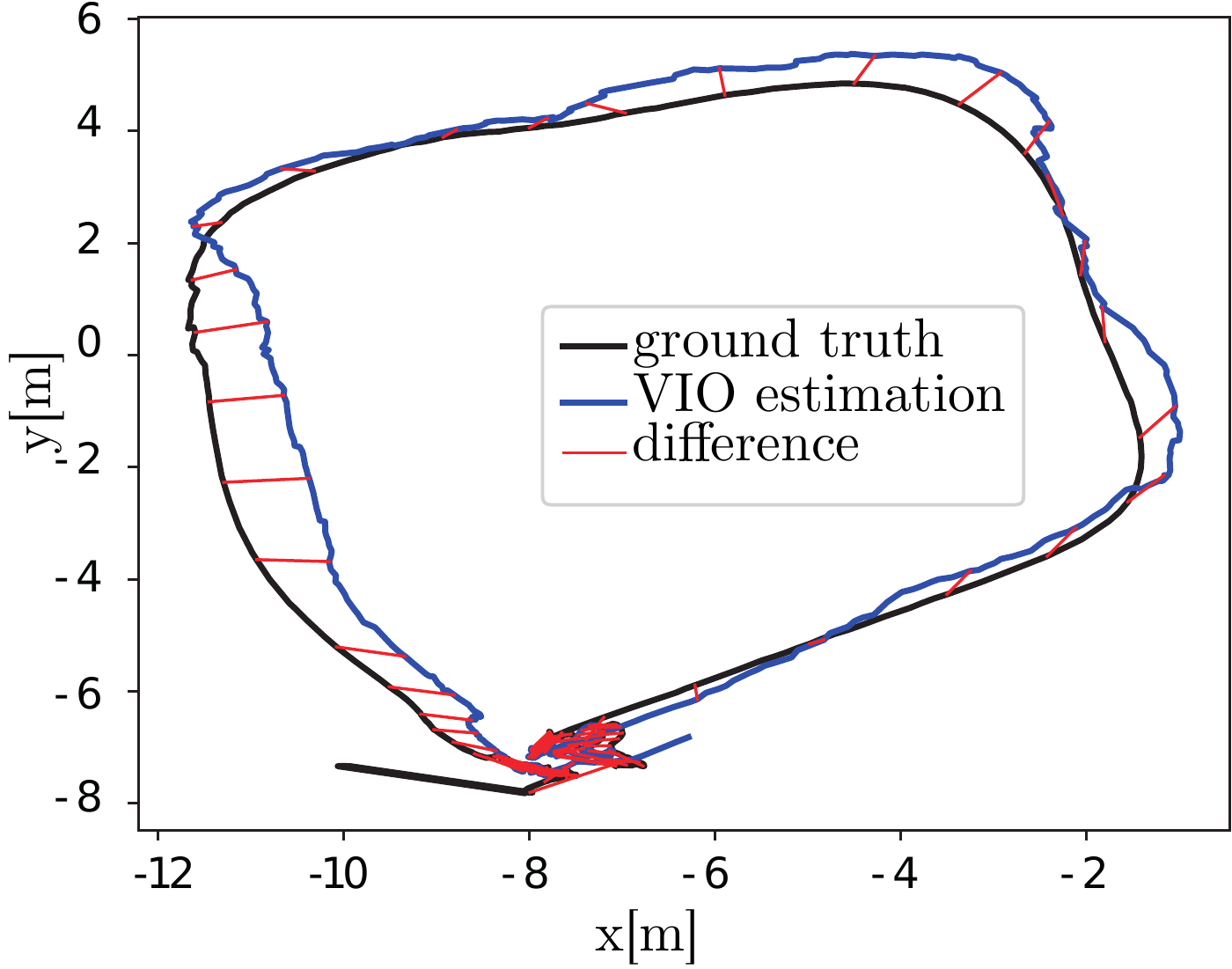}};
    \begin{scope}[x={(a.south east)},y={(a.north west)}]
      \node[fill=black, fill opacity=0.0, text=black, text opacity=1.0] at (0.93, 0.22) {\textbf{(b)}};
      \end{scope}
    \end{tikzpicture}
      \caption[S-MSCKF feedback loop test with fixed UAV orientation]{
       Trajectories flown with S-MSCKF in the feedback of the control system.
        Figure (a) shows trajectory for \SI{0}{\degree} camera orientation, \SI{1}{\meter\per\second} desired velocity with \SI{0.7574}{\meter} ATE.
        Figure (b) shows trajectory for \SI{90}{\degree} camera orientation, \SI{3}{\meter\per\second} desired velocity with \SI{0.4617}{\meter} ATE.
      }
  \label{fig:msckf_real_feedback2}
\end{figure}

\begin{figure}
  \centering
    \begin{tikzpicture}
      \node[anchor=south west,inner sep=0] (a) at (0,0) {\adjincludegraphics[width=0.5\linewidth,trim={{0.00\width} {0.00\height} {0.00\width} {0.00\height}}, clip]{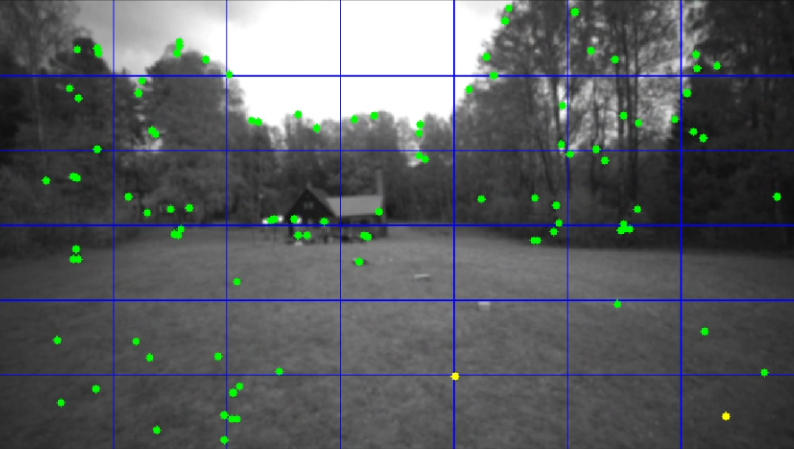}};
    \begin{scope}[x={(a.south east)},y={(a.north west)}]
      \node[fill=white, fill opacity=0.0, text=white, text opacity=1.0] at (0.92, 0.13) {\textbf{(a)}};
      \end{scope}
    \end{tikzpicture}%
    \begin{tikzpicture}
      \node[anchor=south west,inner sep=0] (b) at (0,0) {\adjincludegraphics[width=0.5\linewidth,trim={{0.00\width} {0.00\height} {0.00\width} {0.025\height}}, clip]{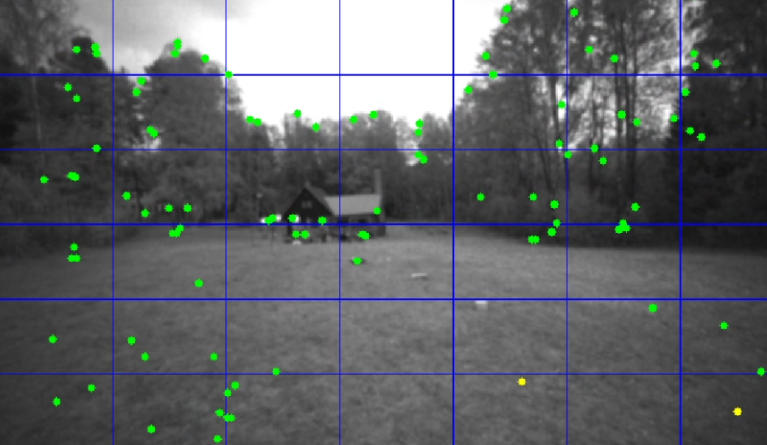}};
    \begin{scope}[x={(a.south east)},y={(a.north west)}]
      \node[fill=white, fill opacity=0.0, text=white, text opacity=1.0] at (0.92, 0.13) {\textbf{(b)}};
      \end{scope}
    \end{tikzpicture}
     \caption[Feedback loop control with SVO algorithm on the real UAV]{
       Left (a) and right (b) camera images with S-MSCKF features during the feedback experiment with \SI{0}{\degree} camera orientation.
  }
  \label{fig:msckf_real_feedback_images}
  \vspace{-1.0em}
\end{figure}

S-MSCKF flight with \SI{0}{\degree} camera orientation is shown in~\autoref{fig:msckf_real_feedback2}a. 
The relatively constant light conditions during the whole flight and minimized camera rapid movements/rotations provide a stable amount of features as shown in~\autoref{fig:msckf_real_feedback_images}.
Surprisingly, the flight with \SI{3}{\meter\per\second} velocity shown in~\autoref{fig:msckf_real_feedback2}b was successful with minimal drift, despite high acceleration values.
Consequently, S-MSCKF might be better on fast flights, as the authors also demonstrated on their custom camera setup.

\section{Conclusions}
\label{sec:conclusion}

This paper provides the first comprehensive study of performance of VIO algorithms under challenging conditions of outdoor deployment of fully autonomous UAVs in high-contrast scenes with direct sunlight.
To achieve a reliable performance of such a lightweight vision system, a trajectory shaping approach based on iterative sampling was proposed to improve the amount of tracked features by reducing the tilting of the UAV.
It was also empirically verified that the negative influence of sunlight is reduced by tilting the camera in the pitch angle, which also increases the number of stable image features.

Selected VIO algorithms were integrated into the feedback of the open-source MRS UAV control system.
All of the tested algorithms provided pose estimates suitable for stable flight in simulation, where the \SI{90}{\degree} camera orientation achieved the best results according to the used metrics.
As expected, with higher UAV velocity, slightly worse results were achieved, but the safe velocity for all tested algorithms is up to \SI{2}{\meter\per\second}, which is sufficient for most intended applications.

In the real-world the algorithms struggled with sunlight and propeller-induced vibrations, which add noise to IMU measurements.
Unfortunately, the VINS-Fusion algorithm was not stable enough to be tested directly in the feedback loop of the real UAV, although it achieved the best results in simulations. 
In the feedback loop of the real UAV control system, the S-MSCKF algorithm achieved the best results from tested algorithms in these challenging conditions.
The advantage of using the S-MSCKF algorithm is that the implementation based on the Kalman filter combines images and IMU messages resulting in satisfying estimation precision during fast movements.
On the other hand, the UAV state is observable only during motion which introduces a position error during hovering.
The SVO algorithm is more sensitive on light conditions because it relies more on captured images than S-MSCKF, but it achieves better performance in slower motions and hovering.


\section*{Acknowledgments}
\label{sec:acknowledgments}
This research was supported by
by CTU grant no SGS20/174/OHK3/3T/13, 
by Technology Agency of the Czech Republic (TACR) project No. FW03010020, 
by the Czech Science Foundation (GAČR) under research project No. 20-29531S, 
by OP VVV funded project CZ.02.1.01/0.0/0.0/16 019/0000765 "Research Center for Informatics", 
by project no. DG18P02OVV069 in program NAKI II, 
by the European Union’s Horizon 2020 research and innovation program AERIAL-CORE under grant agreement no. 871479. 


\balance
\bibliographystyle{IEEEtran}
\bibliography{root}

\end{document}

%% file: Figures/pipeline_diagram.tex
\usetikzlibrary{shapes.geometric,backgrounds,calc}
\pgfdeclarelayer{background}
\pgfdeclarelayer{foreground}
\pgfsetlayers{background,main,foreground}

\tikzset{radiation/.style={{decorate,decoration={expanding waves,angle=90,segment length=4pt}}}}

\begin{tikzpicture}[->,>=stealth', node distance=3.0cm]

  \node[state, shift = {(0.0, 0.0)}] (user_software) {
      \begin{tabular}{c}
        \small Mission \\
        \small planner
      \end{tabular}
    };

  \node[state, right of = user_software, shift = {(0, -0)}] (mpc_tracker) {
      \begin{tabular}{c}
        \small MPC \\
        \small tracker
      \end{tabular}
    };

  \node[state, right of = mpc_tracker, shift = {(0, -0)}] (so3_controller) {
      \begin{tabular}{c}
        \small SE(3) \\
        \small controller
      \end{tabular}
    };

  \node[state, right of = so3_controller, shift = {(0.2, -0)}] (pixhawk) {
      \begin{tabular}{c}
        \small Attitude \\
        \small controller
      \end{tabular}
    };

  \node[state, right of = pixhawk, shift = {(0, -0)}] (uav_plant) {
      \begin{tabular}{c}
        \small UAV \\
        \small plant
      \end{tabular}
    };

  \node[state, below of = so3_controller, shift = {(0, 0.5)}] (state_estimation) {
      \begin{tabular}{c}
        \small State \\
        \small estimation
      \end{tabular}
    };

  \node[state, below of = user_software, shift = {(0, 0.5)}] (stereo_camera) {
      \begin{tabular}{c}
        \small Stereo Camera
      \end{tabular}
    };

  \node[state, below of = state_estimation, shift = {(0, 0.7)}] (vio) {
      \begin{tabular}{c}
        \small VIO
      \end{tabular}
    };

    \path[-] ($(stereo_camera.south) + (0.0, 0)$) edge [] ($(stereo_camera.south |- vio.west)$) -- ($(stereo_camera.south |- vio.west)$) edge [->] node[above, shift = {(0.0, 0.05)}] {  
      \begin{tabular}{c}
        \small images + IMU$^1$
    \end{tabular}}($(vio.west)+(0, 0)$);

    \path[->] ($(vio.north)$) edge [dotted] node[right,midway]{
      \begin{tabular}{c}
        \small VIO pose \\
    \end{tabular}} ($(state_estimation.south)$);

    \path[-] ($(uav_plant.south |- state_estimation.east)$) edge [dotted] ($(uav_plant.south |- vio.east)$) -- ($(uav_plant.south |- vio.east)$) edge [->, dotted] node[above, near end, shift = {(1.4, 0.05)}] {  
      \begin{tabular}{c}
        \small IMU$^2$
    \end{tabular}}($(vio.east)$);

    \path[->] ($(user_software.east) + (0.0, 0)$) edge [] node[above, midway, shift = {(0.0, 0.40)}] {  
      \begin{tabular}{c}
        \small desired \\
        \small trajectory
        \small %
    \end{tabular}} ($(mpc_tracker.west) + (0.0, 0.00)$);

    \path[->] ($(mpc_tracker.east) + (0.0, 0)$) edge [] node[above, midway, shift = {(0.0, 0.40)}] {  
      \begin{tabular}{c}
        \small full-state \\
        \small reference
        \small %
    \end{tabular}} ($(so3_controller.west) + (0.0, 0.00)$);

    \path[->] ($(so3_controller.east) + (0.0, 0)$) edge [] node[above, midway, shift = {(0.0, 0.40)}] {  
      \begin{tabular}{c}
        \small attitude,\\
        \small thrust
        \small %
    \end{tabular}} ($(pixhawk.west) + (0.0, 0.00)$);

    \path[->] ($(pixhawk.east) + (0.0, 0)$) edge [] node[above, midway, shift = {(0.0, 0.40)}] {  
      \begin{tabular}{c}
        \small motor \\
        \small control 
        \small %
    \end{tabular}} ($(uav_plant.west) + (0.0, 0.00)$);

    \pgfmathsetmacro{\offsetA}{-1.5}
    \coordinate (under_mpc) at ($(mpc_tracker.south) + (0.0, \offsetA)$);

    \path[-] ($(state_estimation.west)+(0, 0)$) edge [dotted] node[above, near end, shift = {(0.5, 0.0)}] {  
      \begin{tabular}{c}
        \small UAV state \\
        \small estimate \\
    \end{tabular}} ($(mpc_tracker.south |- state_estimation.west)$) -- ($(mpc_tracker.south |- state_estimation.west)$) edge [->,dotted] ($(mpc_tracker.south)+(0, 0)$);

    \path[->] ($(state_estimation.north)+(0, 0)$) edge [dotted] ($(so3_controller.south)$);

    \path[-] ($(pixhawk.south)+(0, 0)$) edge [dotted] ($(pixhawk.south |- state_estimation.east)$);

    \path[-] ($(uav_plant.south)+(0, 0)$) edge [dotted] ($(uav_plant.south |- state_estimation.east)$) -- ($(uav_plant.south |- state_estimation.east)$) edge [->, dotted] node[below, near start, shift = {(-0.7, 0.0)}] {  
      \begin{tabular}{c}
        \small onboard sensor data \\
        \small %
    \end{tabular}}($(state_estimation.east)$);

  \end{tikzpicture}